\documentclass[11pt]{article}

\usepackage[final]{acl}
\usepackage{subcaption}
\usepackage{float}
\usepackage{times}
\usepackage{xcolor}
\newcommand{\redgain}[1]{\rlap{\textcolor{red}{\tiny\,#1}}}
\usepackage{latexsym}
\usepackage{tabularx}
\usepackage[most]{tcolorbox}
\usepackage{booktabs}
\usepackage{longtable}
\usepackage{array}
\usepackage[T1]{fontenc}

\usepackage[utf8]{inputenc}

\usepackage{microtype}

\usepackage{inconsolata}
\usepackage{placeins}
\usepackage{graphicx}
\usepackage{hyperref}
\usepackage{orcidlink}
\usepackage{caption}
\usepackage{enumitem}
\usepackage{setspace}
\usepackage{ragged2e}

\definecolor{cardborderA}{HTML}{8FA8C9}
\definecolor{cardtitleA}{HTML}{DCE7F5}
\definecolor{cardfillA}{HTML}{F7FAFE}
\definecolor{sectionA}{HTML}{EAF1FB}
\definecolor{evalA}{HTML}{EEF4FC}

\definecolor{cardborderB}{HTML}{A999C8}
\definecolor{cardtitleB}{HTML}{EEE7FA}
\definecolor{cardfillB}{HTML}{FBFAFE}
\definecolor{sectionB}{HTML}{F3EEFC}
\definecolor{evalB}{HTML}{F6F1FD}
\title{VCIFBench: Evaluating Complex Instruction Following for Video Understanding} 

\author{
  \textbf{Huangchen Xu}\textsuperscript{1},
  \textbf{Yuan Wu}\textsuperscript{1,*},
  \textbf{Yi Chang}\textsuperscript{1,2,3}
  \\
  \textsuperscript{1}School of Artificial Intelligence, Jilin University
  \\
  \textsuperscript{2}Engineering Research Center of Knowledge-Driven Human-Machine Intelligence, Jilin University
  \\
  \textsuperscript{3}International Center of Future Science, Jilin University
  \\
  \small{
  \href{mailto:xuhc9924@mails.jlu.edu.cn}{xuhc9924@mails.jlu.edu.cn},
  \href{mailto:yuanwu@jlu.edu.cn}{yuanwu@jlu.edu.cn},
  \href{mailto:yichang@jlu.edu.cn}{yichang@jlu.edu.cn}
  }
  \\
  \small{\textsuperscript{*}Corresponding author.}
}

\begin{document}
\maketitle
\begin{abstract}
Multimodal large language models have made rapid progress in video understanding, yet existing benchmarks largely rely on simple prompts and provide limited evidence about whether models can satisfy explicit output constraints. We introduce VCIFBench, a benchmark for evaluating complex instruction following in video understanding. VCIFBench constructs constraint-rich instructions from both benchmark-adapted and directly video-grounded prompts, covering content, format, style, and structure requirements, and evaluates model outputs with a hybrid verification pipeline. The benchmark contains 306 satisfiable test instructions, 540 DPO training instances, and a 100-item diagnostic set for evaluating whether models can recognize instruction conflicts. Experiments on 10 MLLMs show that joint constraint satisfaction remains challenging. Preference optimization improves instruction following for two model families, while Conflict-100 reveals that models usually execute a satisfiable-looking subset instead of detecting global incompatibility.
\end{abstract}

\begin{figure*}[t]
    \centering
    \includegraphics[width=\linewidth]{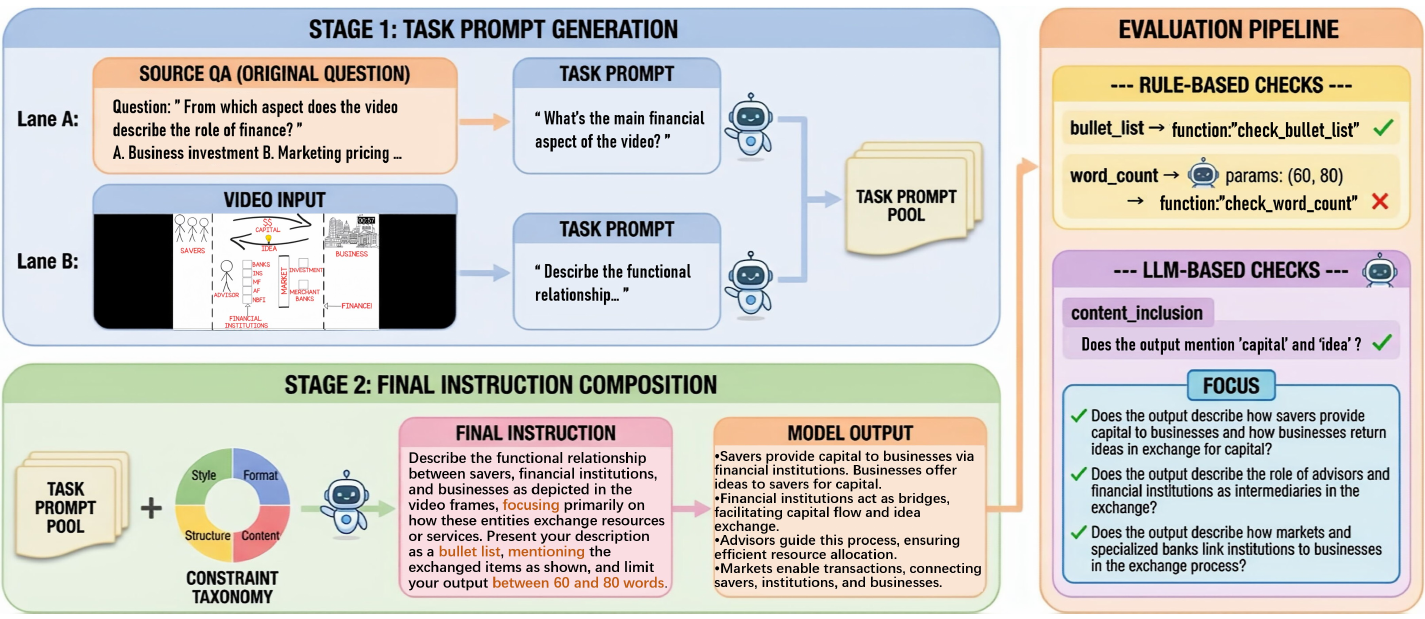}
    \caption{
    Overview of VCIFBench. In Stage 1, we derive task prompts from both original benchmark questions and direct video-grounded prompting. In Stage 2, each final instruction is constructed by combining a task prompt with sampled constraints. The resulting model output is then assessed using a hybrid evaluation pipeline.
    }
    \label{fig:pipeline_overview}
\end{figure*}

\section{Introduction}
\label{sec:intro}

Multimodal large language models (MLLMs) have shown strong capabilities in understanding images, videos, and other heterogeneous inputs \cite{NEURIPS2023_9a6a435e,comanici2025gemini}. A growing set of benchmarks evaluates multimodal reasoning, long-video understanding, and general multimodal competence \cite{yue2024mmmu,zhou2025mlvu,liu2024mmbench,eva}. However, these benchmarks mainly assess task-level performance under relatively simple prompts, leaving unclear whether MLLMs can reliably follow complex instructions with explicit constraints in video-grounded settings.

This gap matters in practical use. A model often needs not only to understand what happens in a video, but also to satisfy user requirements on output content, format, style, and structure. Such requirements are especially important in agentic or tool-use settings, where schema violations or uncontrolled outputs may cause downstream failures \cite{zhang2025appagent}. Although instruction following has been widely studied in language models \cite{he2024can,jiang2024followbench,wen2024benchmarking}, multimodal instruction-following benchmarks remain more limited, often focusing on images \cite{ding2025mm,NEURIPS2023_5503389d} or narrow video tasks such as captioning \cite{li2025if}. Such benchmarks capture only part of the broader task design space. Practical video instructions may also require prediction, temporal or causal reasoning, option selection, or structured generation. As a result, constraints such as length limits, evidence requirements, and ordering rules can behave differently across task types and lead to different failure modes.

Evaluation is another challenge. Rule-based checks are reliable for programmable requirements such as length, format, or exact surface forms \cite{zhou2023instruction}, but they cannot cover many semantic or video-grounded constraints. LLM-based judging is more flexible, but may introduce biases such as length or positional bias \cite{fu2024gptscore,liu-etal-2024-lost}. A useful benchmark therefore needs both diverse constraint coverage and reasonably reliable verification.

We introduce \textbf{VCIFBench}, a benchmark for \textbf{V}ideo \textbf{C}omplex \textbf{I}nstruction \textbf{F}ollowing. VCIFBench constructs constraint-rich instructions from both benchmark-adapted prompts and directly video-grounded prompts, using videos from TempCompass \cite{liu2024tempcompass}, MMWorld \cite{he2024mmworld}, NExT-QA \cite{xiao2021nextqanextphasequestionansweringexplaining}, YouCook2 \cite{zhou2017automaticlearningproceduresweb}, and Video-MME \cite{fu2025video}. The benchmark covers 20 task types and 40 constraint types across content, format, style, and structure. The release includes 306 satisfiable test instructions, 540 DPO training instances, and Conflict-100, a diagnostic of unsatisfiable constraint interactions.

To evaluate model outputs, we develop a hybrid verification pipeline that combines rule-based verification, hybrid parameter extraction with executable checkers, and LLM-based judging. For semantic content constraints, we further decompose high-level requirements into targeted sub-constraints, enabling finer-grained diagnosis of video-grounded failures. Experiments on 10 proprietary and open-source MLLMs show that reliable joint constraint satisfaction remains challenging, especially for video-grounded content constraints. We also show that data derived from the same construction pipeline improve instruction-following performance through DPO on two model families. Code, annotations, and evaluation scripts are available at \url{https://github.com/leaf-ye0749/VCIFBench}.
\section{Related Work}

\subsection{Multimodal Large Language Models and Benchmarks}

Recent multimodal large language models (MLLMs) have demonstrated strong capabilities in understanding and reasoning over images, videos, and other heterogeneous inputs \cite{NEURIPS2023_9a6a435e,comanici2025gemini}. To evaluate these capabilities, many multimodal benchmarks have been proposed, covering settings such as multimodal reasoning \cite{yue2024mmmu}, long-video understanding \cite{zhou2025mlvu}, spatial reasoning \cite{guo2024drivemllm}, and video summarization \cite{jung2025qeva}. These benchmarks have been valuable for tracking model progress and exposing failure modes across tasks and modalities.

Many multimodal benchmarks still emphasize task-level performance under relatively simple prompts, offering limited insight into whether models can satisfy fine-grained requirements on content, format, style, and structure \cite{fu2024mme}. This motivates our focus on complex instruction following in video settings, where task completion must be coupled with explicit constraint satisfaction.

\subsection{Instruction Following}

Instruction following has been studied extensively in language models. Early work focused on explicit and verifiable requirements. For example, IFEval \cite{zhou2023instruction} evaluates programmatically checkable format constraints, while more recent benchmarks study complex instruction following across diverse task types and composition patterns, including combinations of semantic and format constraints \cite{he2024can,wen2024benchmarking}. EifBench further considers settings with multiple instructions and multiple constraints \cite{zou2025eifbench}. Prior work has also explored methods for improving compliance in such scenarios, including in-context learning \cite{zhou2023controlled}, supervised fine-tuning \cite{zhang2025cfbench}, and DPO \cite{ding2025mm}.

In contrast, multimodal instruction following remains less developed, particularly for video. Existing studies mostly examine image-based settings \cite{ding2025mm,NEURIPS2023_5503389d}, while video introduces temporally distributed evidence and more varied response goals. IF-VidCap \cite{li2025if} is an important step toward video-based instruction following, but its scope is mainly caption-oriented. Less is known about how models handle constraints across broader video tasks, such as prediction, option selection, reasoning, and structured response generation. Our work studies this broader setting by combining diverse video task types, multiple constraint categories, and a hybrid evaluation pipeline with rule-based, executable, and judge-based checks.

\section{VCIFBench}

\subsection{Data Collection}

We construct VCIFBench from five complementary video sources, focusing on short videos with durations under two minutes: TempCompass~\cite{liu2024tempcompass}, MMWorld~\cite{he2024mmworld}, NExT-QA~\cite{xiao2021nextqanextphasequestionansweringexplaining}, YouCook2~\cite{zhou2017automaticlearningproceduresweb}, and Video-MME~\cite{fu2025video}. TempCompass supplies temporal and state-change cases; MMWorld covers scientific and general scenarios; NExT-QA contributes causal and temporal reasoning over daily activities; YouCook2 provides procedural cooking; and Video-MME broadens the source pool with diverse real-world activities and domains. The satisfiable test split contains 306 items. The overall construction and evaluation pipeline is shown in Figure~\ref{fig:pipeline_overview}.

\subsection{Instruction Generation}
\label{sec:instruction_generation}

\paragraph{Task taxonomy.}
We organize video-grounded tasks into three broad categories: \textit{perception}, \textit{reasoning}, and \textit{expressive generation}. Unlike benchmarks that mainly evaluate task completion, VCIFBench evaluates whether models can satisfy explicit constraints while performing the task. Detailed task definitions and examples are provided in Appendix~\ref{sec:appendix_task_taxonomy}.

\begin{figure}[t]
    \centering
    \includegraphics[width=0.95\linewidth]{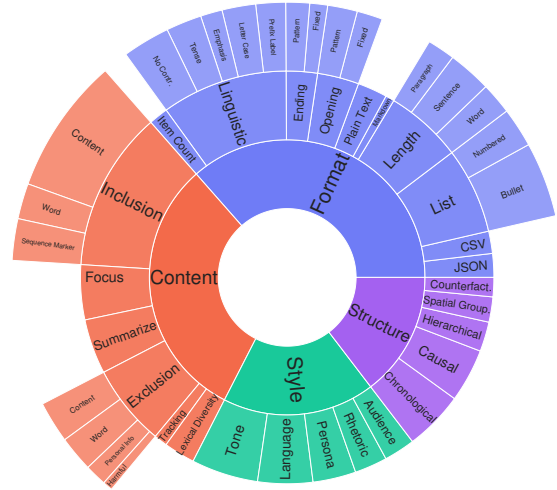}
    \caption{Hierarchical taxonomy of constraints in VCIFBench.}
    \label{fig:constraint_taxonomy}
\end{figure}

\paragraph{Constraint taxonomy.}
We build on prior taxonomies of practical instruction constraints~\cite{jiang2024followbench,zou2025eifbench,li2025if} and adapt them to video instruction following. As shown in Figure~\ref{fig:constraint_taxonomy}, VCIFBench contains 40 constraint types organized into four dimensions. \textbf{Format} constraints specify surface requirements such as serialization, length, and fixed patterns. \textbf{Content} constraints require models to select, include, exclude, or emphasize video-supported information. \textbf{Style} constraints control communicative form, such as tone, target audience, persona, and rhetoric. \textbf{Structure} constraints specify organization requirements, such as chronological, causal, or hierarchical ordering. Importantly, although content-related constraints depend on video understanding, faithfulness to the video content is itself a core aspect of instruction following, as a helpful response must remain grounded in the video rather than rely on unsupported assumptions.

\paragraph{Compositional instruction types.}
Following prior work~\cite{wen2024benchmarking}, we construct instructions with several composition patterns: \textbf{And} for jointly satisfying multiple constraints, \textbf{Chain} for combining subtasks, \textbf{Selection} for conditioning the response on explicit alternatives, and \textbf{Nested} for hierarchical combinations. These descriptive tags are not mutually exclusive---for example, \textbf{And} commonly appears inside \textbf{Chain} and \textbf{Selection} instructions.

\paragraph{Instruction construction.}
We construct instructions from two complementary sources, using GPT-5.2~\cite{openai2025gpt52} as the primary generator.

\textbf{Benchmark-derived prompts.}
For source datasets with QA-style task descriptions, we rewrite suitable prompts into open-ended video tasks and combine them with compatible constraints. We discard cases that cannot be meaningfully transformed, such as questions asking about content absent from the video. TempCompass also provides original--reverse video pairs, which naturally support binary \textbf{Selection} instructions and allow us to test whether models follow video-grounded alternatives rather than passing a single branch by chance.

\textbf{Direct video-grounded prompts.}
For videos without suitable annotations, we select tasks compatible with observable video evidence and then compose them with candidate constraints. GPT-5.2 generates and revises the instruction to improve fluency. We further use Gemini-2.5 Pro~\cite{comanici2025gemini} to check whether content-related constraints are supported by the visible video, followed by manual revision to correct residual inconsistencies. Appendix~\ref{sec:appendix_base306} reports the full constraints map.

\subsection{Evaluation}
\label{sec:evaluation}

VCIFBench uses a hybrid evaluation pipeline because constraints differ in how directly they can be verified.

\paragraph{Rule-based verification.}
For constraints with deterministic criteria, such as JSON validity or required surface forms, we directly implement checking functions and compute binary outcomes.

\paragraph{Hybrid verification.}
Some constraints are rule-checkable only after their parameters are identified from the instruction. 
For example, a word-count constraint requires extracting the target range before applying a checker. 
Following prior work on constrained and tool-use evaluation~\cite{zhong2025complexfuncbench,wen2024benchmarking}, we ask the judge model to extract the required parameters from the instruction and then execute the corresponding checker.

\paragraph{LLM-based judging.}
For constraints that depend on semantic interpretation, discourse organization, or video-grounded reasoning, we use GPT-5~\cite{singh2025openai} to produce binary constraint-level judgments, following prior work on LLM-as-a-judge, multimodal LLM-based evaluation, and complex instruction evaluation~\cite{zheng2023judging,yu2024mm,zou2025eifbench,wen2024benchmarking}. 
To make video-grounded evaluation more reliable, we further decompose content-related constraints, such as summarization, focus, tracking, and content inclusion, into explicit evidence-grounded sub-constraints. We report judge--human alignment in Appendix~\ref{sec:appendix_judge_alignment}.

\paragraph{Metrics.}
VCIFBench is constraint-centric. We report both constraint-level and instruction-level metrics. Let $\mathcal{S}$ be the set of samples. For each sample $i \in \mathcal{S}$, let $\mathcal{C}_i$ denote its annotated constraints, and let $p_{i,c} \in \{0,1\}$ indicate whether constraint $c$ is satisfied.

\textbf{CPR (Constraint Pass Rate).}
For a set of annotated constraints $\mathcal{A}$, we define
\[
\mathrm{CPR}(\mathcal{A})=
\frac{1}{|\mathcal{A}|}\sum_{(i,c)\in\mathcal{A}} p_{i,c}.
\]
$\mathcal{A}$ can correspond to all constraints of a given type or dimension.

\textbf{IPR (Instruction Pass Rate).}
IPR measures whether all constraints in an instruction are satisfied jointly:
\[
\mathrm{IPR}
=\frac{1}{|\mathcal{S}|}\sum_{i\in\mathcal{S}}
\mathbf{1}\!\left[\forall c\in\mathcal{C}_i,\; p_{i,c}=1\right].
\]
This all-constraints criterion reflects practical settings where violating one critical requirement can invalidate the output.

\textbf{Conflict diagnostic evaluation.}
Conflict-100 is excluded from standard IPR/CPR aggregation because its instructions are intentionally unsatisfiable. Models receive only the video and a natural instruction, without any cue that a conflict may exist. We separately measure whether the response explicitly detects incompatibility and whether it completely localizes every requirement in the annotated minimal unsatisfiable core.

\subsection{DPO Dataset Construction}
\label{sec:dpo_dataset_construction}

We construct 540 underlying DPO training instances from the same instruction-construction pipeline. Each instance is represented by two chosen--rejected comparisons.
Chosen responses are generated by GPT-5.2 and then processed through the same evaluation pipeline and manual revision to ensure constraint satisfaction and video grounding.

For every underlying instance, the two comparisons jointly cover complementary negative-construction routes. Inspired by MM-IFEngine's MM-IFDPO negative-pair construction~\cite{ding2025mm}, one route removes the explicit constraints and asks MiniCPM-o-4.5 to answer the underlying video task, producing a plausible but constraint-insensitive response. The other prompts MiniCPM-o-4.5 to corrupt a specific constraint in the chosen response while preserving the rest of the answer as much as possible.
We manually revise samples from the second strategy where MiniCPM-o-4.5 only introduces superficial changes without creating a clear instruction-following violation.
\section{Experiments}

\subsection{Experimental Setup}

Our final satisfiable test set contains 306 samples. We evaluate 10 multimodal large language models (MLLMs): Gemini-3-Pro \cite{gemini3modelcard}, Gemini-3-Flash \cite{gemini3flashmodelcard}, GPT-5.4 \cite{openai2026gpt54}, Qwen3-VL-30B-A3B \cite{bai2025qwen3vltechnicalreport}, Qwen3-VL-8B \cite{bai2025qwen3vltechnicalreport}, InternVL3.5-14B \cite{wang2025internvl3}, InternVL3.5-8B \cite{wang2025internvl3}, MiMo-VL-7B-RL \cite{mimovl2025technicalreport}, MiniCPM-o-4.5 \cite{yao2024minicpm}, and GLM-4.1V-9B-Thinking \cite{hong2025glm}. Table~\ref{tab:base306_main_failure_severity} reports instruction pass rate (IPR), constraint pass rate (CPR), and failure severity on Base-306.

We evaluate 10 MLLMs under the same input setting for the main results. 
All videos are uniformly sampled at 0.5 fps and capped at 1080p, with the generation temperature set to 0.1.

\subsection{Main Results}
\label{sec:main_results}

Table~\ref{tab:base306_main_failure_severity} reports the Base-306 results. 
The leading proprietary models form a close group, with IPR around 52\% but CPR around 85--87\%. 
This gap shows that satisfying individual constraints does not reliably translate into satisfying the full instruction: errors accumulate when all constraints must be satisfied jointly. 
Across dimensions, Style and Structure are relatively strong for proprietary models, while Content remains lower at around 70--71\%. 
Format is also not saturated, indicating that even explicit surface requirements, such as word-count limits and JSON or CSV formatting, can still fail under multi-constraint settings.
\begin{table}[t]
\centering
\scriptsize
\setlength{\tabcolsep}{2.2pt}
\renewcommand{\arraystretch}{1.02}
\caption{
Base-306 main results with failure severity. 
Near denotes failed instructions with exactly one violated constraint; Viol. is the average number of violated constraints per failed instruction.
}
\label{tab:base306_main_failure_severity}
\resizebox{\columnwidth}{!}{%
\begin{tabular}{@{}lrrrrrrrr@{}}
\toprule
\textbf{Model} & \textbf{IPR} & \textbf{CPR} & \textbf{Fmt.} & \textbf{Cont.} & \textbf{Sty.} & \textbf{Str.} & \textbf{Near} & \textbf{Viol.} \\
\midrule
Gemini-3-Pro   & 52.94 & 86.59 & 91.39 & 71.00 & 95.10 & 92.86 & 75.69 & 1.438 \\
GPT-5.4        & 51.96 & 87.18 & 92.49 & 70.77 & 94.77 & 95.24 & 74.83 & 1.347 \\
Gemini-3-Flash & 51.63 & 85.69 & 89.36 & 70.07 & 95.75 & 93.45 & 66.89 & 1.493 \\
\midrule
MiniCPM-o-4.5  & 29.41 & 74.03 & 79.81 & 55.92 & 85.95 & 76.79 & 48.15 & 1.856 \\
Qwen3-VL-30B-A3B & 28.43 & 73.38 & 81.53 & 54.99 & 83.01 & 72.02 & 50.23 & 1.877 \\
Qwen3-VL-8B    & 27.12 & 74.94 & 81.38 & 57.31 & 85.29 & 76.79 & 53.81 & 1.735 \\
GLM-4.1V-9B-Thinking & 26.80 & 75.06 & 81.38 & 56.84 & 86.27 & 77.38 & 52.68 & 1.719 \\
InternVL3.5-14B & 24.84 & 68.91 & 78.72 & 51.04 & 73.20 & 69.64 & 43.04 & 2.087 \\
MiMo-VL-7B-RL  & 24.51 & 71.57 & 82.00 & 52.44 & 78.76 & 67.86 & 45.02 & 1.900 \\
InternVL3.5-8B &  5.56 & 46.37 & 47.10 & 40.60 & 51.31 & 49.40 & 22.49 & 2.865 \\
\bottomrule
\end{tabular}%
}
\end{table}
Open-source models show substantially lower instruction-level success. 
Most achieve CPR around 69--74\%, but their IPR remains only around 24--29\%, suggesting that they can satisfy many individual constraints but are less stable when constraints are composed. 
InternVL3.5-8B is the clearest outlier, with low scores across all dimensions.

Since IPR only indicates whether an instruction fails, we further report near-miss rate and the average number of violated constraints per failed instruction. 
The severity metrics show that failures are not homogeneous. 
For Gemini-3-Pro and GPT-5.4, more than 73\% of failed instructions are near misses, and each failed instruction violates only 1.44 and 1.35 constraints on average. 
By contrast, mid-tier open-source models have lower near-miss rates and higher violation counts. 
Thus, stronger models often remain close to satisfying all requirements, whereas weaker models more often exhibit cascading failures across multiple constraints.

Manual inspection supports this distinction. 
Gemini and GPT failures are usually narrow boundary errors: the semantic answer is often coherent and video-grounded, but one explicit constraint boundary is violated, such as a word budget, a CSV/JSON rule, a missing visual detail, or an excluded extra detail. 
Some semantic judgments are inherently more interpretation-sensitive, so strong models may be closer to the practical ceiling on these cases. 
However, many remaining failures are still concrete compliance errors, suggesting that multi-constraint instructions place additional pressure even on otherwise simple constraints. 
In contrast, InternVL3.5-8B failures more often indicate global breakdowns, including mixed languages, answering both branches of a selection instruction, or uncontrolled long continuations.
\paragraph{Option-order robustness.}
We also test selection robustness by swapping the display order of Option A and Option B. 
This diagnostic asks whether models follow the video-grounded branch condition, rather than relying on the surface position or label of an option. 
As shown in Table~\ref{tab:option_order_robustness}, Gemini-3-Pro and GPT-5.4 are much more stable under option-order swaps. 
By contrast, Qwen3-VL-8B and MiniCPM-o-4.5 show much lower robustness: they sometimes select the correct path under one option order, but fail to preserve the same decision after the options are swapped. 
This indicates that \texttt{selection\_accuracy} is not merely about choosing the correct branch once; it also requires resisting superficial option-order effects and maintaining a stable video-grounded decision.
\begin{table}[t]
\centering
\scriptsize
\setlength{\tabcolsep}{3.5pt}
\renewcommand{\arraystretch}{0.98}
\caption{Option-order robustness on 26 selection pairs.}
\label{tab:option_order_robustness}
\begin{tabular}{@{}lcc@{}}
\toprule
\textbf{Model} & \textbf{Both Corr.} & \textbf{Either Corr.} \\
\midrule
Gemini-3-Pro & 21/26 & 23/26 \\
GPT-5.4 & 17/26 & 21/26 \\
Qwen3-VL-8B & 4/26 & 11/26 \\
MiniCPM-o-4.5 & 2/26 & 7/26 \\
\bottomrule
\end{tabular}
\end{table}
\subsection{Effect of Constraint Count}
\label{sec:constraint_count}

\begin{figure}[t]
    \centering
    \includegraphics[width=0.95\linewidth]{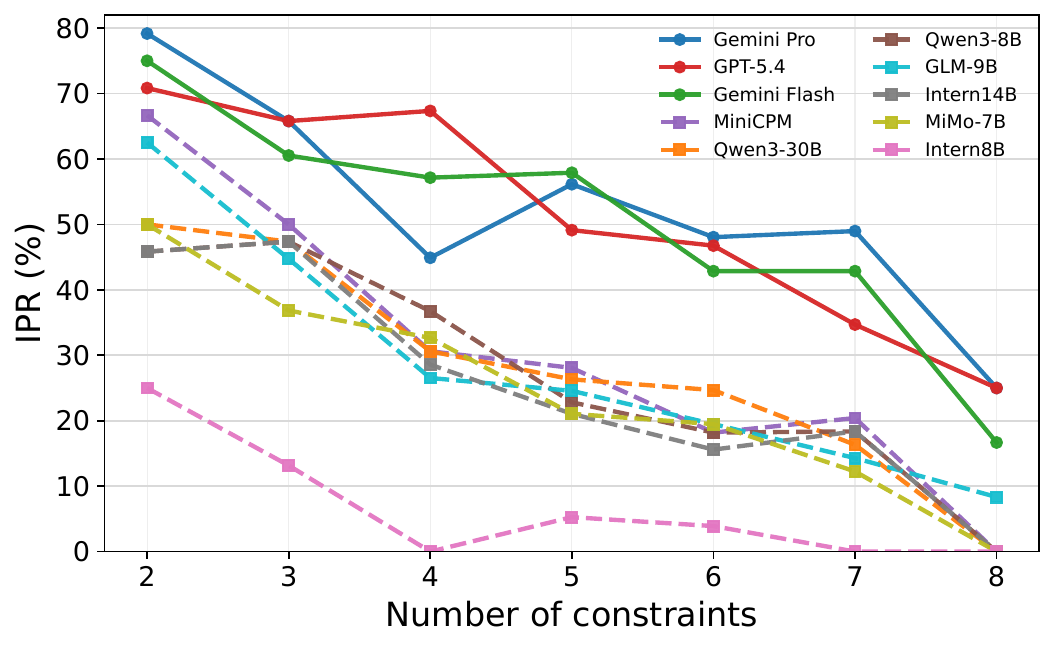}
    \caption{Instruction pass rate by number of constraints on Base-306 across ten selected models. The aggregate trend decreases as the number of constraints increases.}
    \label{fig:ipr_by_constraint_count}
\end{figure}

We group Base-306 samples by the number of annotated constraints in each instruction, denoted as $k=|\mathcal{C}_i|$, and compute IPR over sample--run pairs across the ten selected models. As shown in Figure~\ref{fig:ipr_by_constraint_count}, the aggregate trend decreases as $k$ increases: IPR drops from 57.08\% at $k=2$ to 7.50\% at $k=8$. The high-$k$ buckets should be interpreted cautiously, since $k=8$ contains only 12 unique samples.

This trend supports the view that instruction-level failures accumulate under multi-constraint settings. However, constraint count alone does not determine difficulty. Some model-level curves show local rebounds because different $k$ buckets contain different types of constraints. For example, Gemini-3-Pro improves from $k=4$ to $k=5$, as the $k=4$ bucket contains more difficult video-grounded content and selection constraints, with failures concentrated in \texttt{content\_inclusion}, \texttt{summarize} and \texttt{selection\_accuracy}. By contrast, the $k=5$ bucket includes more format-, style-, and list-oriented constraints, such as \texttt{plain\_text}, \texttt{bullet\_list}, and \texttt{tone}, which this model satisfies more reliably. Similar rebounds appear for MiniCPM-o-4.5 and InternVL3.5-14B.

Overall, the results suggest that both constraint count and constraint composition matter: instructions with more constraints are generally harder to satisfy, but local variations depend on which constraint types appear together.

\subsection{Constraint Failure Mechanisms}
\label{sec:constraint_failure_mechanisms}

We further analyze why models fail on Base-306. Rather than treating all failed instructions as the same type of error, we distinguish between constraints that are difficult in themselves and constraints that become fragile mainly when composed with other requirements. This distinction helps explain why relatively high constraint-level performance does not necessarily translate into high instruction-level success.

\paragraph{From hard constraints to failure mechanisms.}
The hardest constraints in Base-306 come from several sources. Some require video-grounded evidence selection and semantic compression, such as \texttt{summarize}, \texttt{content\_inclusion}, and \texttt{focus}. Others involve negative or boundary control, such as \texttt{content\_exclusion} and \texttt{lexical\_diversity}, where the model must avoid prohibited or repeated information while still producing a useful response. We also find that seemingly simple surface requirements, such as \texttt{word\_count} and \texttt{csv}, can remain fragile, especially for weaker models.

\begin{table}[t]
\centering
\footnotesize
\setlength{\tabcolsep}{3.2pt}
\renewcommand{\arraystretch}{0.98}
\caption{
Representative intrinsic and interaction-sensitive constraints. 
Low-mix denotes instructions with at most three constraints, and high-mix denotes instructions with at least six constraints.
}
\label{tab:intrinsic_interaction_constraints}
\begin{tabular}{lrrr}
\toprule
\textbf{Constraint} & \textbf{Low-mix} & \textbf{High-mix} & \textbf{Drop} \\
\midrule
\multicolumn{4}{c}{\textit{Intrinsically difficult}} \\
\midrule
\texttt{summarize} & 21.67 & 30.00 & -8.33 \\
\texttt{content\_exclusion} & 35.00 & 53.33 & -18.33 \\
\texttt{content\_inclusion} & 57.24 & 46.73 & 10.51 \\
\texttt{lexical\_diversity} & 44.00 & 60.00 & -16.00 \\
\midrule
\multicolumn{4}{c}{\textit{Interaction-sensitive}} \\
\midrule
\texttt{pattern\_closing} & 87.50 & 60.00 & 27.50 \\
\texttt{csv} & 80.00 & 61.43 & 18.57 \\
\texttt{sentence\_count} & 90.00 & 70.00 & 20.00 \\
\texttt{output\_language} & 87.50 & 73.67 & 13.83 \\
\bottomrule
\end{tabular}
\end{table}

\paragraph{Interaction patterns.}
We further examine whether some co-occurring constraints fail more often than expected from their individual failure rates. This analysis is diagnostic, since constraint types are not randomly assigned and some pairs have limited unique samples. Still, manual inspection suggests several recurring patterns.

\begin{itemize}[leftmargin=1.2em,itemsep=0.35em,topsep=0.25em]

    \item \textbf{Length/count budget $\times$ required content or structure.}
    Length and count constraints become fragile when paired with content or structure requirements. Models often exceed the budget to include visual evidence, causal links, or required markers; conversely, they may satisfy the budget by omitting required visual anchors. This pattern suggests a trade-off between being complete and staying within strict output limits.

    \item \textbf{Selection branch.}
    \texttt{selection\_accuracy} can act as a gatekeeper constraint. Once a model chooses the wrong branch, preserves both branches, or leaves option labels in the answer, subsequent constraints often fail as a cascading consequence. 

    \item \textbf{Style or rhetoric $\times$ rigid surface format.}
    Persona, tone, and rhetorical constraints can encourage greetings, labels, explanatory framing, or narrative openings. These additions may improve naturalness, but they can violate exact openings, fixed closings, plain-text requirements, or label placement. In creative writing or role-play tasks, models may produce more readable responses while breaking exact surface constraints.

    \item \textbf{Output language/rhetoric $\times$ lexical exclusion or diversity.}
    A model may avoid the exact prohibited word while reintroducing the excluded content through paraphrase, adjacent concepts, or plausible elaboration, especially when instructed to produce the response in another language.
\end{itemize}

Overall, Base-306 failures should not be interpreted as isolated constraint errors. Some constraints are intrinsically difficult because they require video-grounded evidence selection or semantic compression. Others become difficult mainly under composition, where content, format, style, and task-specific response tendencies compete with one another.

\subsection{Effect of Visual Budget}
\label{sec:visual_budget}

\begin{figure}[t]
    \centering
    \includegraphics[width=\columnwidth]{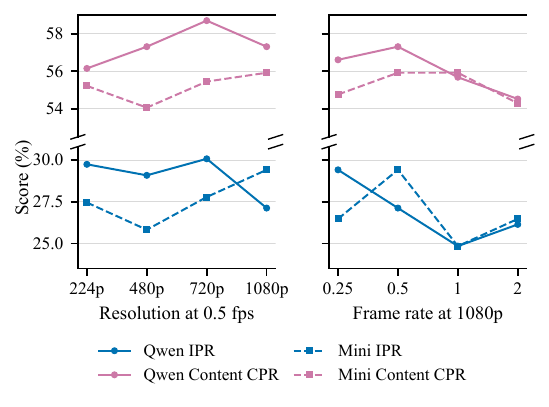}
    \caption{
    Visual-budget trends on Base-306 for Qwen3-VL-8B and MiniCPM-o-4.5. The left panel varies the maximum-side spatial cap at 0.5 fps, and the right panel varies the frame rate under the 1080p max-side setting. Solid lines report Content CPR, while dashed lines report IPR.
    }
    \label{fig:visual_budget_overall_trend}
\end{figure}

We further examine whether larger visual input budgets improve complex video instruction following. The spatial labels 224p, 480p, 720p, and 1080p denote maximum-side caps after resizing. Table~\ref{tab:qwen8_visual_budget_compact} reports the Qwen3-VL-8B ablation, while Figure~\ref{fig:visual_budget_overall_trend} compares Qwen3-VL-8B and MiniCPM-o-4.5.

\begin{table}[t]
\centering
\small
\setlength{\tabcolsep}{4.5pt}
\renewcommand{\arraystretch}{1.05}
\caption{
Qwen3-VL-8B visual-budget ablation on Base-306. CPR is micro-averaged over all constraint instances, and Content denotes Content CPR.
}
\label{tab:qwen8_visual_budget_compact}
\resizebox{\columnwidth}{!}{%
\begin{tabular}{lrrrrr}
\toprule
\textbf{Setting} & \textbf{IPR} & \textbf{CPR} & \textbf{Content} & \textbf{Tokens} & \textbf{Words} \\
\midrule
224p, 0.5 fps   & 29.74 & 74.48 & 56.15 & 1841  & 234.5 \\
480p, 0.5 fps   & 29.08 & 73.70 & 57.31 & 3268  & 218.6 \\
720p, 0.5 fps   & \textbf{30.07} & 75.00 & \textbf{58.70} & 6463  & 217.4 \\
1080p, 0.25 fps & 29.41 & \textbf{75.65} & 56.61 & 5845  & 164.0 \\
1080p, 0.5 fps  & 27.12 & 74.94 & 57.31 & 11402 & 207.4 \\
1080p, 1 fps    & 24.84 & 73.51 & 55.68 & 22519 & 224.8 \\
1080p, 2 fps    & 26.14 & 73.64 & 54.52 & 37519 & 268.8 \\
\bottomrule
\end{tabular}%
}
\end{table}

The results show that larger visual budgets are not monotonically beneficial. For Qwen3-VL-8B, 720p at 0.5 fps yields the highest IPR and Content CPR, whereas 1080p at 0.25 fps yields the highest overall CPR. Increasing the spatial cap to 1080p or using higher frame rates substantially increases the input-token budget without improving full-instruction success. MiniCPM-o-4.5 exhibits a different optimum, achieving its highest IPR and CPR at 1080p and 0.5 fps, followed by lower performance at 1 fps and 2 fps.

Content compliance also varies non-monotonically and does not track instruction-level success exactly. Qwen3-VL-8B reaches its highest Content CPR at 720p and 0.5 fps, while MiniCPM-o-4.5 reaches its highest Content CPR at 1080p and either 0.5 or 1 fps. Overall, additional visual tokens can alter content compliance, but they do not by themselves guarantee that all instruction constraints will be satisfied jointly.

\subsection{Conflict-Aware Instruction Following}
\label{sec:conflict_aware_instruction_following}

Conflict-100 evaluates whether models check the joint satisfiability of a natural video instruction before attempting to execute it. In practical applications, users may inadvertently combine incompatible constraints because of input errors or iterative editing. A reliable model should detect such incompatibility and request clarification, rather than silently satisfying a convenient subset of the instruction.

The benchmark contains 100 unique videos and ten incompatibility mechanisms, comprising 70 binary and 30 ternary minimal conflicting constraint sets. Each constituent constraint is individually reasonable, and removing any constraint from a conflicting set makes the remaining requirements jointly satisfiable. For example, a cooking instruction may require the response to mention visible ingredients while simultaneously requiring every substantive noun phrase to refer only to a person or utensil. Further construction and family-level details are given in Appendix~\ref{sec:conflict_diagnostic_subset}.

\begin{figure}[t]
    \centering
    \includegraphics[width=\columnwidth]{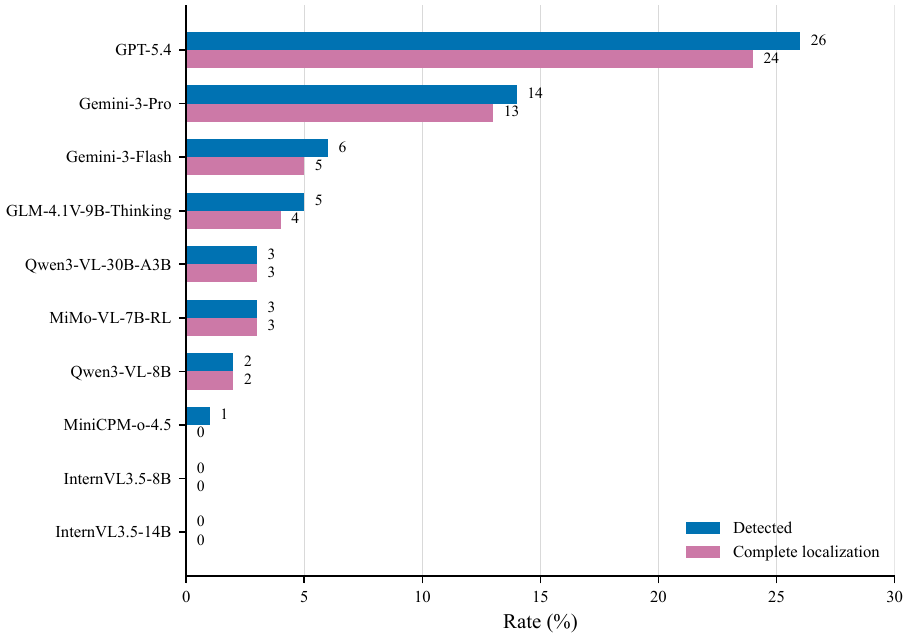}
    \caption{Conflict-100 model-level results (\%). \textit{Detected} requires an explicit statement that the instruction's requirements cannot all be satisfied; \textit{Complete localization} additionally requires identifying every conflict in the minimal conflicting set. }
    \label{fig:conflict100_model_funnel}
\end{figure}

Figure~\ref{fig:conflict100_model_funnel} shows that only 60 of 1,000 model--item responses detect a conflict and 54 completely localize it. Moreover, 65 of 100 items are never detected by any model, and complete localization is at most 8\% in every conflict family.

To focus specifically on conflict reasoning, we inspect 152 well formed and visually grounded responses. Of these responses, 94.1\% attempt to execute the task without recognizing the incompatible constraints, while only 5.9\% show conflict awareness. Eight conflict aware responses propose a repair, and one refuses to proceed. The dominant failure is selective compliance, with 60.5\% satisfying a convenient subset of the instruction while overlooking or violating other constraints. Among applicable cases with clear preference evidence, 101 of 104 responses prioritize completing the base task over checking satisfiability, and 23 of 25 follow positive requirements over prohibitions. These results show that even when generation and video understanding are successful, models tend to sacrifice less salient constraints to produce an answer instead of first determining whether the complete instruction is jointly satisfiable.

\subsection{Preference Optimization Results}
\label{sec:dpo_results}

We apply LoRA-based DPO to Qwen3-VL-8B and MiniCPM-o-4.5 to examine whether the preference signal transfers across model families.  Table~\ref{tab:dpo_qwen8_results} reports instruction- and constraint-level performance for both backbones.

\begin{table}[t]
\centering
\scriptsize
\setlength{\tabcolsep}{2.2pt}
\renewcommand{\arraystretch}{1.02}
\caption{
DPO results for two model families on Base-306.
}
\label{tab:dpo_qwen8_results}
\resizebox{\columnwidth}{!}{%
\begin{tabular}{l r@{\hspace{2.8em}} r@{\hspace{2.8em}} rrrrr}
\toprule
\textbf{Run} & \textbf{IPR} & \textbf{CPR-micro} & \textbf{CPR-macro} & \textbf{Fmt.} & \textbf{Cont.} & \textbf{Sty.} & \textbf{Str.} \\
\midrule
Qwen3-VL-8B Base & 27.12 & 74.94 & 75.76 & 81.38 & 57.31 & 85.29 & 76.79 \\
Qwen3-VL-8B DPO  & 33.01\redgain{(+5.88)} & 78.63\redgain{(+3.69)} & 80.40 & 84.82 & 60.56 & 90.85 & 79.17 \\
MiniCPM-o-4.5 Base & 29.41 & 74.03 & 76.29 & 79.81 & 55.92 & 85.95 & 76.79 \\
MiniCPM-o-4.5 DPO & 36.93\redgain{(+7.52)} & 77.85\redgain{(+3.82)} & 79.68 & 84.98 & 58.24 & 86.60 & 85.12 \\
\bottomrule
\end{tabular}%
}
\end{table}

DPO consistently improves both backbones on Base-306. Qwen3-VL-8B gains 5.88 points in IPR and 3.69 points in CPR-micro, while MiniCPM-o-4.5 gains 7.52 and 3.82 points, respectively. Both models also improve in CPR-macro and across all four constraint dimensions, although their largest gains occur in different dimensions. These results demonstrate that the preference signal transfers across model families. Appendix~\ref{sec:appendix_external_mmifeval} reports the corresponding transfer evaluation on MM-IFEval~\cite{ding2025mm}.

\section{Conclusion}

We present \textbf{VCIFBench}, a benchmark for evaluating complex instruction following in video understanding, together with a construction pipeline and a hybrid evaluation pipeline. Experiments on 10 MLLMs show that even capable models struggle to satisfy multiple explicit constraints jointly. Conflict-100 further reveals a strong task-completion bias when locally plausible requirements are globally incompatible. Preference optimization improves Base-306 performance for two model families and transfers modestly to MM-IFEval. We hope that VCIFBench supports the development of more reliable and controllable multimodal systems.

\section{Limitations}

First, our current benchmark primarily centers on English instructions and outputs. Broader multilingual coverage would be a useful extension for studying complex video instruction following across languages and linguistic conventions.

Second, VCIFBench intentionally covers a focused subset of video instruction following. 
Consequently, VCIFBench should be viewed as a controlled stress test for constraint-rich video instruction following. 
Its results are informative about whether models can jointly satisfy explicit requirements, but should not be used as a standalone measure of overall video understanding, factual grounding, or deployment safety.

Finally, our evaluation pipeline combines rule-based verification, hybrid verification, and LLM-based judging to balance coverage and reliability. Although this design improves practical coverage, variations in judge behavior and interpretation cannot be fully ruled out, especially for more semantic or subjective constraints.

\paragraph{Potential Risks.}
VCIFBench is a controlled evaluation resource and does not introduce direct deployment risks. A possible indirect risk is that optimizing models primarily for explicit constraint satisfaction could encourage formal compliance at the expense of broader reliability, such as factual grounding, safety, or appropriate refusal behavior. We therefore recommend using VCIFBench alongside broader evaluations of video understanding, factuality, and safety.

\section*{Acknowledgments}

This work was supported by the National Key Research and Development Program of China under Grant No.~2023YFF0905400, the National Natural Science Foundation of China under Grant No.~U2341229, and the Reform Commission Foundation of Jilin Province under Grant No.~2024C003.

\bibliography{custom}

\appendix
\section{Additional Details}
\label{sec:appendix_details}

\subsection{Additional Implementation and Dataset Details}

All local experiments were conducted on NVIDIA RTX PRO 6000 GPUs with 96 GB of memory. 
Each DPO training run took approximately two hours.
For inference, InternVL3.5-14B and Qwen3-VL-30B-A3B used two GPUs, while the remaining local model runs used a single GPU. 
The total local inference and evaluation time across all experiments was approximately 32 hours.

We use the five source benchmarks solely for noncommercial academic research and retain their original source identifiers. TempCompass is released under CC BY-NC 4.0. MMWorld provides its dataset under CC BY 4.0. NExT-QA releases its code and question answering resources under the MIT License, while YouCook2 distributes its annotations under the MIT License. The Video-MME benchmark is available for academic research use. We only release our newly constructed annotations. Copyright in all source videos remains with their original creators.

\subsection{Base-306 Constraint Composition}
\label{sec:appendix_base306}

\begin{figure}[t]
    \centering
    \includegraphics[width=\columnwidth]{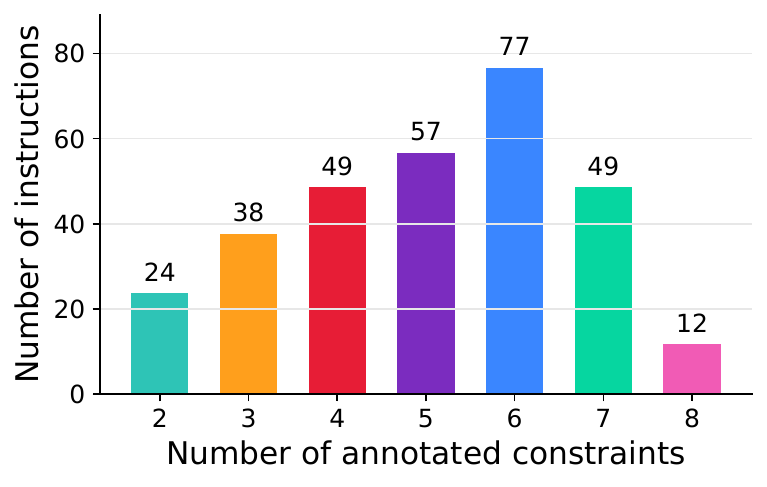}
    \caption{Number of annotated constraints per instruction in Base-306.}
    \label{fig:appendix_constraint_count_distribution}
\end{figure}

Figure~\ref{fig:appendix_constraint_count_distribution} shows the distribution of the number of constraints per satisfiable instruction in Base-306. Most items contain four to seven constraints, which makes instruction-level success substantially stricter than satisfying any single constraint in isolation.

\begin{figure*}[p]
    \centering
    \includegraphics[width=\textwidth]{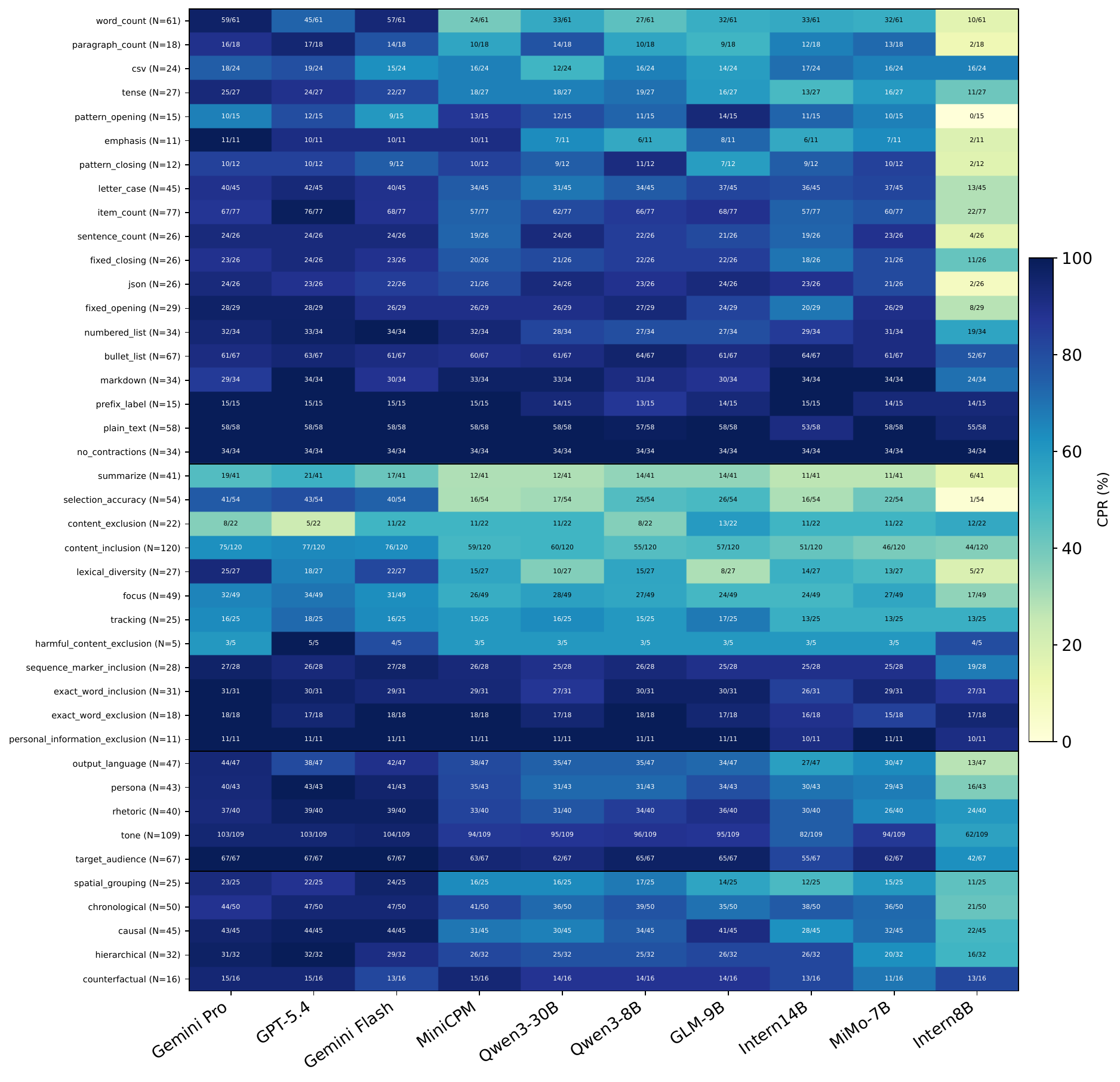}
    \caption{Full Base-306 constraint-type heatmap. The y-axis lists constraint types with the number of annotated instances in parentheses. Cell annotations report passed/annotated-total model attempts, and color indicates CPR.}
    \label{fig:appendix_base_constraint_heatmap}
\end{figure*}

Figure~\ref{fig:appendix_base_constraint_heatmap} reports the full constraint-type heatmap. Rows are constraint types, grouped by dimension and sorted by average CPR within each group. Cell text gives the number of passed model attempts over the annotated total for that constraint type. Empty outputs and API-blocked outputs are counted as failures: in our runs, two Gemini test items were blocked by the API safety filter, and five InternVL3.5-8B items produced empty outputs.

\subsection{Conflict Diagnostic Subset}
\label{sec:conflict_diagnostic_subset}

\paragraph{Construction and validation.}
We use separate GPT-5.6 instances \cite{openai2026gpt56} to construct and review each item based on its source video and original task. The construction instance drafts and revises the instruction and its conflict annotation so that each constraint in a binary or ternary set is individually satisfiable and only the complete annotated set is jointly unsatisfiable. The review instance then checks video accuracy, scene compatibility, proper subset witnesses, alternative interpretations, formatting and scope escapes. Items that fail any check are revised or replaced and reviewed again.

\begin{figure}[t]
    \centering
    \includegraphics[width=\columnwidth]{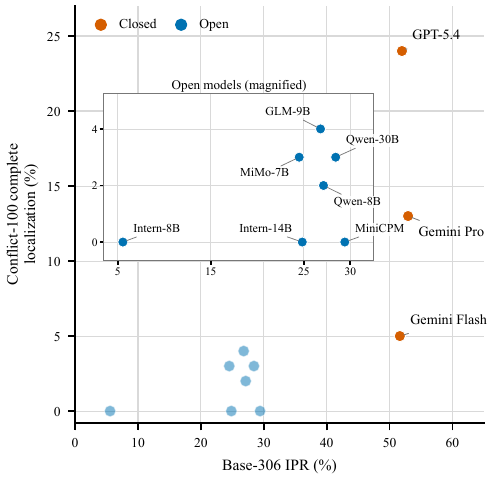}
    \caption{Base-306 instruction pass rate versus Conflict-100 complete localization.}
    \label{fig:base_ipr_conflict100_localization}
\end{figure}

\begin{figure*}[t]
    \centering
    \includegraphics[width=\textwidth]{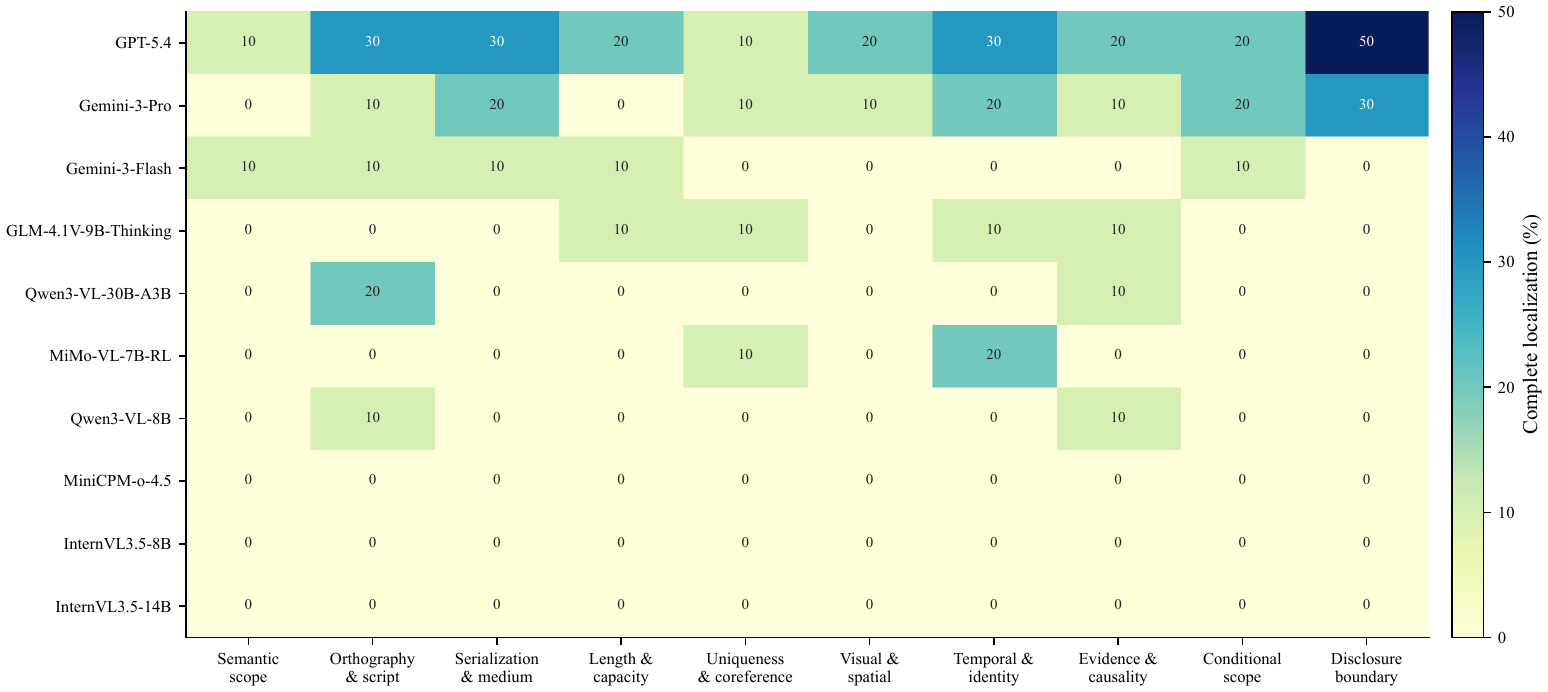}
    \caption{Complete localization percentage across the ten Conflict-100 incompatibility mechanisms.}
    \label{fig:conflict100_family_heatmap}
\end{figure*}

Table~\ref{tab:conflict_diagnostic_templates} summarizes the ten incompatibility mechanisms and their construction operators. Figure~\ref{fig:conflict100_model_funnel} reports detection and complete localization by model. Figure~\ref{fig:base_ipr_conflict100_localization} compares complete localization with Base-306 instruction pass rate, while Figure~\ref{fig:conflict100_family_heatmap} reports localization across incompatibility mechanisms. The complete evaluator prompt is provided in Figure~\ref{fig:conflict_judge_prompt}.

\begin{table*}[t]
\centering
\scriptsize
\setlength{\tabcolsep}{3.1pt}
\renewcommand{\arraystretch}{1.0}
\caption{Conflict-100 families, construction operators, and aggregate results across ten models.}
\label{tab:conflict_diagnostic_templates}
\begin{tabularx}{\textwidth}{p{3.2cm}Xrr}
\toprule
\textbf{Conflict family} & \textbf{Construction operator} & \textbf{Detect} & \textbf{Localize} \\
\midrule
Semantic scope and entailment & Require a concrete member while excluding its semantic class or attribute domain. & 2\% & 2\% \\
Orthography, case, and script & Combine exact transcription with incompatible case, alphabet, or normalization properties. & 10\% & 8\% \\
Serialization and output medium & Require representation properties for tree ownership, shared identity, or medium that cannot coexist. & 6\% & 6\% \\
Length, count, and capacity & Make local lower bounds exceed a global word, field, slot, or sequence capacity. & 4\% & 4\% \\
Repetition, uniqueness, and coreference & Combine stable identity, unique naming, and coreference relations that violate equality or transitivity. & 5\% & 4\% \\
Static visual and spatial grounding & Pair grounded entities with a forced near-miss attribute, count, hierarchy, or spatial relation. & 3\% & 3\% \\
Temporal state and cross-frame identity & Compose before/after states, event order, or cross-frame identity into an impossible trajectory. & 9\% & 8\% \\
Evidence, causality, and counterfactual & Restrict evidence to visible facts while demanding an exact hidden cause, secret, prediction, or counterfactual. & 7\% & 6\% \\
Conditional selection and scope & Use video-triggered branches and nested scopes whose activated requirements cannot all hold. & 5\% & 5\% \\
Disclosure and information governance & Require a specific identifier or topic while excluding the semantic or information class containing it. & 9\% & 8\% \\
\bottomrule
\end{tabularx}
\end{table*}

\subsection{Residual Failure Modes in Instruction Construction}

After automated verification, we manually inspect residual inconsistencies in the satisfiable split. Despite multi-stage LLM-based filtering, several recurring failure modes remain.

First, some instructions contain mismatches between content-related constraints and visual evidence, for example, by requiring details or events that do not appear in the video. We also observe occasional accidental cross-constraint conflicts, such as exact-word requirements that are incompatible with the required output language. These accidental cases are removed or repaired in the satisfiable split; by contrast, the separate conflict diagnostic subset in Appendix~\ref{sec:conflict_diagnostic_subset} deliberately contains hard conflicts for diagnostic evaluation.

Second, some \textbf{Selection}-type instructions remain underspecified. In these cases, the unselected branch is not described clearly enough, or the branching condition is framed too subjectively (e.g., whether the response should be written ``as a news report'' or ``as a tutorial''), rather than being tied to an objectively identifiable alternative in the video. This weakens the intended design of \textbf{Selection}, where the correct branch should be uniquely determined by visual evidence.

\subsection{Human Validation and Judge Alignment}
\label{sec:appendix_judge_alignment}

To assess the reliability of our semantic evaluation pipeline, we conducted an additional human validation study with three undergraduate annotators from computer science and related disciplines who had no prior experience with VCIFBench. Before annotation, they received task-specific training and a codebook defining every evaluated constraint type, its satisfaction criteria, and positive and negative examples for important decision boundaries. They independently labeled 630 constraint-level instances and were compensated at USD 9 per hour. Post-annotation quality control rechecked decisions on objective requirements, including count limits and output-language constraints, and found no systematic careless-labeling pattern. The annotators reached 96.19\% complete agreement, with Fleiss' $\kappa = 0.9342$; their majority vote is the human reference for judge-alignment analysis.

Because our main judge uses a closed-source model, we further examine whether an alternative judge model can serve as a more reproducible option for deployment or independent replication. 
Table~\ref{tab:appendix_judge_human_alignment} compares GPT-5 and GLM-4.5V against the human majority vote. 
GPT-5 shows stronger alignment with the human reference, reaching 88.89\% agreement and $\kappa$ = 0.7300. 
GLM-4.5V obtains lower but still moderate alignment, with 84.92\% agreement. 

\begin{table}[H]
\centering
\scriptsize
\setlength{\tabcolsep}{3.2pt}
\renewcommand{\arraystretch}{1.02}
\caption{
Judge alignment with the human majority reference. 
}
\label{tab:appendix_judge_human_alignment}
\resizebox{\columnwidth}{!}{%
\begin{tabular}{lrrrrr}
\toprule
\textbf{Comparison} & \textbf{$n$} & \textbf{Agreement} & \textbf{Kappa} & \textbf{AC1} & \textbf{PABAK} \\
\midrule
GPT-5 vs. human & 630 & 88.89 & 0.7300 & 0.8114 & 0.7778 \\
GLM-4.5V vs. human & 630 & 84.92 & 0.5555 & 0.7742 & 0.6984 \\
\bottomrule
\end{tabular}%
}
\end{table}
\subsection{Task--Constraint Interaction Diagnostics}
\label{sec:appendix_task_constraint_interaction}

The main text summarizes recurring constraint--constraint interactions. 
Here we provide additional task--constraint diagnostics, focusing on cases where a constraint becomes harder because the task itself encourages a competing response style. 

\begin{itemize}[leftmargin=1.2em,itemsep=0.3em,topsep=0.25em]

    \item \textbf{State-change detection $\times$ tone.}
    Although \texttt{tone} is generally not a hard constraint, it becomes more fragile when the task asks for state changes. 
    In one sample, models are asked to describe a visible facial change neutrally rather than interpret the emotion, but outputs such as ``he appears to be happy'' or ``broad, happy smile'' turn visual change into affective interpretation. 
    This suggests that task semantics can pull models away from neutral description and make style control harder.

    \item \textbf{\texttt{practical\_writing} $\times$ \texttt{rhetoric}.}
    Practical writing favors direct, actionable steps, while rhetorical constraints require comparison, contrast, or simile. 
    In YouCook2 examples, some models produce clear cooking instructions but omit the required rhetorical device, while others add figurative language that is not clearly grounded in the visible action. 
    This reflects a tension between practical clarity and rhetorical elaboration.

\end{itemize}
\subsection{Detailed Task Taxonomy}
\label{sec:appendix_task_taxonomy}

Table~\ref{tab:task_glossary} summarizes the task taxonomy used in VCIFBench. The definitions are taken from the task descriptions used in prompts during instruction construction.

\subsection{Detailed Constraint Taxonomy}
\label{sec:appendix_constraint_taxonomy}

Table~\ref{tab:constraint_glossary} summarizes the constraint taxonomy used in VCIFBench. 
We construct the taxonomy through a bottom-up and then top-down procedure. 
We first collect candidate constraints from prior instruction-following benchmarks, video-grounded tasks, and the requirements that naturally arise when adapting video questions into open-ended instructions. 
We then group these candidates according to the aspect of the response they primarily control.

This leads to a distinction between \textit{what is said} and \textit{how it is expressed}. 
Constraints about required, excluded, or emphasized video-supported information are grouped under \textbf{Content}. 
The ``how'' side is divided into \textbf{Format}, covering mechanical surface requirements such as serialization, casing, and length; \textbf{Style}, covering communicative expression such as tone, persona, target audience, and output language; and \textbf{Structure}, covering information organization such as chronological, causal, hierarchical, and spatial grouping. 

\begin{table*}[!t]
\centering
\small
\setlength{\tabcolsep}{6pt}
\renewcommand{\arraystretch}{1.18}
\caption{Detailed glossary of task types in VCIFBench, including concise definitions.}
\label{tab:task_glossary}
\begin{tabularx}{\textwidth}{p{4.15cm}X}
\toprule
\textbf{Task Type} & \textbf{Definition} \\
\midrule

\multicolumn{2}{c}{\textbf{Perception}} \\
\midrule
emotion\_recognition & Infer emotions only when they are clearly visible from facial expression or body language. \\
harmful\_content\_detection & Detect harmful or unsafe content only when supported by explicit visual evidence. \\
text\_recognition & Extract readable on-screen text only when it is clearly visible. \\
\midrule

\multicolumn{2}{c}{\textbf{Reasoning}} \\
\midrule
event\_sequencing & Determine the temporal order of events in the video. \\
state\_change\_detection & Describe what changes from earlier to later stages based on visual evidence. \\
temporal\_query & Answer before, during, or after questions with justification grounded in the frames. \\
spatial\_arrangement\_description & Describe the layout of objects or entities using relative spatial positions such as left, right, inside, or behind. \\
object\_relationship & Describe physical or functional relationships among objects, such as support, attachment, containment, or blocking. \\
cause\_effect\_identification & Identify cause--effect relations only when they are visually supported, avoiding unsupported speculation. \\
counterfactual\_reasoning & Reason about what would likely change under an alternative visual condition. \\
predictive\_reasoning & Predict likely outcomes only when the prediction is strongly grounded in the observed visual context. \\
\midrule

\multicolumn{2}{c}{\textbf{Expressive Generation}} \\
\midrule
summarization & Produce a faithful high-level summary of the video rather than a transcript. \\
tutorial\_generation & Convert what is shown in the video into step-by-step instructions. \\
professional\_writing & Present the content in a formal style, such as a report, briefing, or meeting minutes. \\
practical\_writing & Express the content in an actionable form, such as a checklist, plan, or troubleshooting guide. \\
platform\_copywriting & Produce platform-oriented copy grounded in the facts shown in the video. \\
role\_play\_response & Respond consistently from a specified persona while remaining grounded in the video content. \\
creative\_writing & Generate creative text that does not contradict the visible facts. \\
style\_transfer\_rewriting & Rewrite the content in a specified style while preserving factual consistency. \\
point\_of\_view\_reframing & Reframe the same content from a different viewpoint without changing the underlying facts. \\
\bottomrule
\end{tabularx}
\end{table*}

\begin{table*}[!t]
\centering
\small
\setlength{\tabcolsep}{5pt}
\renewcommand{\arraystretch}{1.10}
\caption{Detailed glossary of constraint types in VCIFBench, including their definitions and evaluation types.}
\label{tab:constraint_glossary}
\begin{tabularx}{\textwidth}{
>{\raggedright\arraybackslash}p{4.1cm}
>{\raggedright\arraybackslash}p{2.0cm}
X}
\toprule
\textbf{Constraint Type} & \textbf{Evaluation} & \textbf{Definition} \\
\midrule

\multicolumn{3}{c}{\textbf{Format}} \\
\midrule
json & Rule-based & Output must be valid JSON (object or array). \\
csv & Rule-based & Output must be valid CSV. \\
bullet\_list & Rule-based & Use an unordered bullet list. \\
numbered\_list & Rule-based & Use an ordered list. \\
markdown & Rule-based & Use markdown formatting. \\
plain\_text & Rule-based & Plain text only. \\

word\_count & Hybrid & Enforce word count bounds. \\
sentence\_count & Hybrid & Enforce sentence count bounds. \\
paragraph\_count & Hybrid & Enforce paragraph count bounds. \\
item\_count & Hybrid & Enforce item or list entry count bounds. \\

fixed\_opening & Hybrid & Start with the exact specified string. \\
fixed\_closing & Hybrid & End with the exact specified string. \\
pattern\_opening & LLM-based & The opening should follow a described pattern rather than an exact fixed string. \\
pattern\_closing & LLM-based & The closing should follow a described pattern rather than an exact fixed string. \\
prefix\_label & Hybrid & Each item should begin with the specified prefix or label format. \\

letter\_case & Hybrid & Apply the required casing rule to the specified scope. \\
emphasis & LLM-based & Emphasize specified spans with the given markers. \\
tense & LLM-based & Use the specified tense consistently. \\
no\_contractions & Rule-based & Avoid contractions (e.g., use ``do not'' instead of ``don't''). \\

\midrule
\multicolumn{3}{c}{\textbf{Content}} \\
\midrule
summarize & LLM-based & Require abstraction or a high-level summary rather than literal transcription. \\
focus & LLM-based & Emphasize the specified aspects and allocate most of the content to them. \\

content\_inclusion & LLM-based & The response must mention specified visible items, which should remain small in scope and checkable. \\
exact\_word\_inclusion & Hybrid & The response must include the specified exact word or phrase verbatim. \\
sequence\_marker\_inclusion & LLM-based & The response must use specified ordering or connective markers. \\
tracking & LLM-based & Re-identify recurring entities across scenes while maintaining cross-scene identity consistency. \\
lexical\_diversity & Hybrid & Avoid repeating key terms; this may specify banned terms or a maximum repetition count. \\

content\_exclusion & LLM-based & The response must not mention specified items. \\
exact\_word\_exclusion & Hybrid & The response must not include the specified exact word or phrase. \\
personal\_information\_exclusion & LLM-based & Do not reveal personal information, only when people or identity-bearing details are visible. \\
harmful\_content\_exclusion & LLM-based & Do not include harmful or unsafe content. \\

\midrule
\multicolumn{3}{c}{\textbf{Style}} \\
\midrule
tone & LLM-based & Use the specified tone. \\
target\_audience & LLM-based & Adapt the response for the specified audience. \\
persona & LLM-based & Write from the specified role or persona. \\
output\_language & LLM-based & Produce the entire response in the specified language or script. \\
rhetoric & LLM-based & Use the specified rhetorical devices. \\

\midrule
\multicolumn{3}{c}{\textbf{Structure}} \\
\midrule
chronological & LLM-based & Organize the response in real-time chronological order. \\
causal & LLM-based & Present cause before effect with explicit linking. \\
spatial\_grouping & LLM-based & Organize the response according to spatial layout. \\
hierarchical & LLM-based & Organize the response by abstraction level. \\
counterfactual & LLM-based & Organize the response into three parts: observed facts, intervention, and predicted outcome. \\
\bottomrule
\end{tabularx}
\end{table*}
\subsection{External Evaluation of DPO-Tuned Model}
\label{sec:appendix_external_mmifeval}
\begin{table}[H]
\centering
\scriptsize
\setlength{\tabcolsep}{4pt}
\renewcommand{\arraystretch}{1.02}
\caption{
External evaluation on MM-IFEval. 
}
\label{tab:external_mmifeval_results}
\begin{tabular*}{\columnwidth}{@{\extracolsep{\fill}}l r@{\hspace{2.4em}} r@{\hspace{2.4em}} r@{\hspace{2.4em}}@{}}
\toprule
\textbf{Model} & \textbf{P-level} & \textbf{C-level} & \textbf{Overall} \\
\midrule
Qwen3-VL-8B Base & 47.00 & 75.20 & 68.15 \\
Qwen3-VL-8B DPO  & 48.00\redgain{(+1.00)} & 77.02\redgain{(+1.81)} & 69.76\redgain{(+1.61)} \\
MiniCPM-o-4.5 Base & 31.00 & 66.95 & 57.96 \\
MiniCPM-o-4.5 DPO & 29.00 & 70.50\redgain{(+3.55)} & 60.12\redgain{(+2.16)} \\
\bottomrule
\end{tabular*}
\end{table}
To examine transfer beyond VCIFBench, we evaluate both trained backbones on MM-IFEval~\cite{ding2025mm}. C-level measures output-composition compliance, whereas P-level emphasizes perception tied to the visual input; Overall is the item-weighted mean over 300 C-level and 100 P-level instances. Qwen3-VL-8B improves from 68.15 to 69.76, with gains at both levels. MiniCPM-o-4.5 improves from 57.96 to 60.12.
\section{Examples}
\label{sec:ex}

In this section, we provide representative examples generated and evaluated by our pipeline, including instructions constructed by our generation pipeline, corresponding model outputs, evaluation results produced by our evaluation pipeline, and the prompts used in our implementation.

\clearpage


\tcbset{
  exampleouter/.style={
    breakable,
    enhanced jigsaw,
    colback=cardfillA,
    colframe=cardborderA,
    boxrule=1pt,
    arc=3mm,
    width=0.94\textwidth,
    left=3mm,right=3mm,top=3mm,bottom=3mm
  },
  exampletitle/.style={
    colback=cardtitleA,
    colframe=cardtitleA,
    boxrule=0pt,
    arc=2mm,
    left=2mm,right=2mm,top=1.2mm,bottom=1.2mm
  },
  examplesection/.style={
    breakable,
    enhanced jigsaw,
    colback=sectionA,
    colframe=cardborderA,
    boxrule=0.6pt,
    arc=1.8mm,
    fonttitle=\bfseries,
    fontupper=\small\justifying,
    left=2mm,right=2mm,top=1.5mm,bottom=1.5mm
  },
  checksection/.style={
    breakable,
    enhanced jigsaw,
    colback=evalA,
    colframe=cardborderA,
    boxrule=0.7pt,
    arc=1.8mm,
    fonttitle=\bfseries,
    fontupper=\small\justifying,
    left=2mm,right=2mm,top=1.5mm,bottom=1.5mm
  },
  constraintbox/.style={
    breakable,
    enhanced jigsaw,
    colback=white,
    colframe=cardborderA,
    boxrule=0.7pt,
    arc=1.8mm,
    left=1.8mm,right=1.8mm,top=1.5mm,bottom=1.5mm,
    fontupper=\small\justifying,
    coltitle=black,
    fonttitle=\bfseries\small,
    colbacktitle=cardtitleA!70,
    title=#1
  },
  evalitem/.style={
    breakable,
    enhanced jigsaw,
    colback=sectionA!45!white,
    colframe=cardborderA,
    boxrule=0.45pt,
    arc=1.2mm,
    left=1.5mm,right=1.5mm,top=1mm,bottom=1mm,
    fontupper=\small\justifying
  }
}

\newcommand{\Pass}{\textbf{\textcolor{green!50!black}{Pass}}}
\newcommand{\Fail}{\textbf{\textcolor{red!70!black}{Fail}}}
\newcommand{\YesAns}{\textbf{\textcolor{green!40!black}{Yes}}}
\newcommand{\NoAns}{\textbf{\textcolor{red!70!black}{No}}}

\clearpage

\begin{figure*}[p]
\centering
\begin{tcolorbox}[exampleouter, breakable]

\begin{tcolorbox}[exampletitle]
\textbf{\LARGE Example 1.}
\end{tcolorbox}

\vspace{1mm}

\begin{center}
\fcolorbox{cardborderA}{white}{
    \includegraphics[width=0.90\linewidth,height=4cm,keepaspectratio]{figs/example1.pdf}
}
\end{center}

\vspace{1mm}

\begin{tcolorbox}[examplesection, title=\textbf{Instruction}]
Describe the transformation of the plant from earlier to later frames based on the video, choosing between two options: \textbf{Option A} (the plant is reviving) and \textbf{Option B} (the plant is wilting).

If the plant is reviving, describe the observed transformation of the plant from earlier frames to later frames, focusing on the overall improvement trend across frames and using the present continuous tense consistently when describing the frames.

If the plant is wilting, describe how the plant changes from earlier to later frames, focusing on the overall decline trend across frames, ensure that no descriptive word about the flower's condition is repeated more than twice, and emphasize visible signs of deterioration by marking them with double asterisks (**).
\end{tcolorbox}

\vspace{1.2mm}

\begin{tcolorbox}[examplesection, colback=white, title=\textbf{Model Output}]
\textbf{Option A: The plant is reviving.} The leaves, which earlier appeared drooping and limp, are now gradually lifting and standing more upright. The overall posture of the plant is improving, with each frame showing a subtle but consistent upward movement of the foliage. The greenery appears to be gaining vigor, with the edges of the leaves becoming less curled and more taut. The plant’s stems are visibly straightening, and the canopy is expanding slightly, indicating renewed energy. The transformation continues with the leaves regaining their natural shape, and the plant’s presence becomes more vibrant and lively across the frames.
\end{tcolorbox}

\vspace{1.2mm}

\begin{tcolorbox}[checksection, title=\textbf{LLM-BASED CHECKS}]

\begin{tcolorbox}[constraintbox={\texttt{focus}\hfill \Pass}]
\begin{tcolorbox}[evalitem]
\textbf{Question 1:} Does the output describe an overall improvement trend, specifically noting that the leaves are gradually lifting, consistent with the plant reviving?

\vspace{0.35em}
\textbf{Result:} \Pass
\end{tcolorbox}
\end{tcolorbox}

\vspace{0.7em}

\begin{tcolorbox}[constraintbox={\texttt{tense}\hfill \Pass}]
\begin{tcolorbox}[evalitem]
\textbf{Constraint:} Use present continuous tense consistently when describing the frames.

\vspace{0.35em}
\textbf{Result:} \Pass
\end{tcolorbox}
\end{tcolorbox}

\vspace{0.7em}

\begin{tcolorbox}[constraintbox={\texttt{selection\_accuracy}\hfill \Pass}]
\begin{tcolorbox}[evalitem]
\textbf{Question 1:} Does the model select the correct branch consistent with: \emph{The plant is reviving}?

\vspace{0.35em}
\textbf{Result:} \Pass
\end{tcolorbox}
\end{tcolorbox}

\end{tcolorbox}

\end{tcolorbox}

\label{fig:appendix_example_reviving}

\end{figure*}

\clearpage

\begin{figure*}[p]
\centering

\begin{tcolorbox}[exampleouter]

\begin{tcolorbox}[exampletitle]
\textbf{\LARGE Example 2.}
\end{tcolorbox}

\vspace{1mm}

\begin{center}
\fcolorbox{cardborderA}{white}{
    \includegraphics[width=0.90\linewidth,height=4cm,keepaspectratio]{figs/example2.pdf}
}
\end{center}

\vspace{1mm}

\begin{tcolorbox}[examplesection, title=\textbf{Instruction}]
Describe the transformation of the plant from earlier to later frames based on the video, choosing between two options: \textbf{Option A} (the plant is reviving) and \textbf{Option B} (the plant is wilting).

If the plant is reviving, describe the observed transformation of the plant from earlier frames to later frames, focusing on the overall improvement trend across frames and using the present continuous tense consistently when describing the frames.

If the plant is wilting, describe how the plant changes from earlier to later frames, focusing on the overall decline trend across frames, ensure that no descriptive word about the flower's condition is repeated more than twice, and emphasize visible signs of deterioration by marking them with double asterisks (**).
\end{tcolorbox}

\vspace{1mm}

\begin{tcolorbox}[examplesection, colback=white, title=\textbf{Model Output}]
\textbf{Option B: The plant is wilting.} Across the frames, its leaves are visibly drooping and sagging, with some edges curling inward. The overall posture of the plant is increasingly limp, and the leaves appear less vibrant, with a noticeable loss of turgidity. \textbf{The stems are bending downward}, and \textbf{the foliage is losing its upright stance}. The transformation indicates a steady decline in vitality, with no signs of recovery or improvement.
\end{tcolorbox}

\vspace{1.2mm}

\begin{tcolorbox}[checksection, title=\textbf{HYBRID CHECKS}]

\begin{tcolorbox}[constraintbox={\texttt{lexical\_diversity}\hfill \Pass}]
\begin{tcolorbox}[evalitem]
\textbf{Function:}

{\ttfamily\small
check\_lexical\_diversity(model\_output, words, max\_repeats)
}

\vspace{0.3em}
\textbf{Extracted parameters:}
\begin{itemize}[leftmargin=1.5em, itemsep=0.15em, topsep=0.2em]
    \item \texttt{words = [wilting, drooping, sagging, limp, vibrant, decline, deterioration]}
    \item \texttt{max\_repeats = 2}
\end{itemize}

\vspace{0.2em}
\textbf{Result:} \Pass
\end{tcolorbox}
\end{tcolorbox}

\end{tcolorbox}

\vspace{1mm}

\begin{tcolorbox}[checksection, title=\textbf{LLM-BASED CHECKS}]

\begin{tcolorbox}[constraintbox={\texttt{emphasis}\hfill \Pass}]
\begin{tcolorbox}[evalitem]
\textbf{Constraint:} Emphasize visible signs of deterioration by marking them with double asterisks.

\vspace{0.3em}
\textbf{Result:} \Pass
\end{tcolorbox}
\end{tcolorbox}

\end{tcolorbox}

\end{tcolorbox}
\caption{A representative example of selection.}
\label{wilt}
\end{figure*}
\clearpage
\begin{figure*}[p]
\centering

\begin{tcolorbox}[exampleouter]

\begin{tcolorbox}[constraintbox={\texttt{selection\_accuracy}\hfill \Pass}]
\begin{tcolorbox}[evalitem]
\textbf{Question 1:} Does the output select the correct branch consistent with: \emph{The plant is wilting}?

\vspace{0.3em}
\textbf{Result:} \Pass
\end{tcolorbox}
\end{tcolorbox}

\end{tcolorbox}
\end{figure*}
\vspace{1.2mm}
\begin{figure*}[p]
\centering

\begin{tcolorbox}[exampleouter]

\begin{tcolorbox}[exampletitle]
\textbf{\LARGE Example 3.}
\end{tcolorbox}

\vspace{1mm}

\begin{center}
\fcolorbox{cardborderA}{white}{
    \includegraphics[width=0.90\linewidth,height=4cm,keepaspectratio]{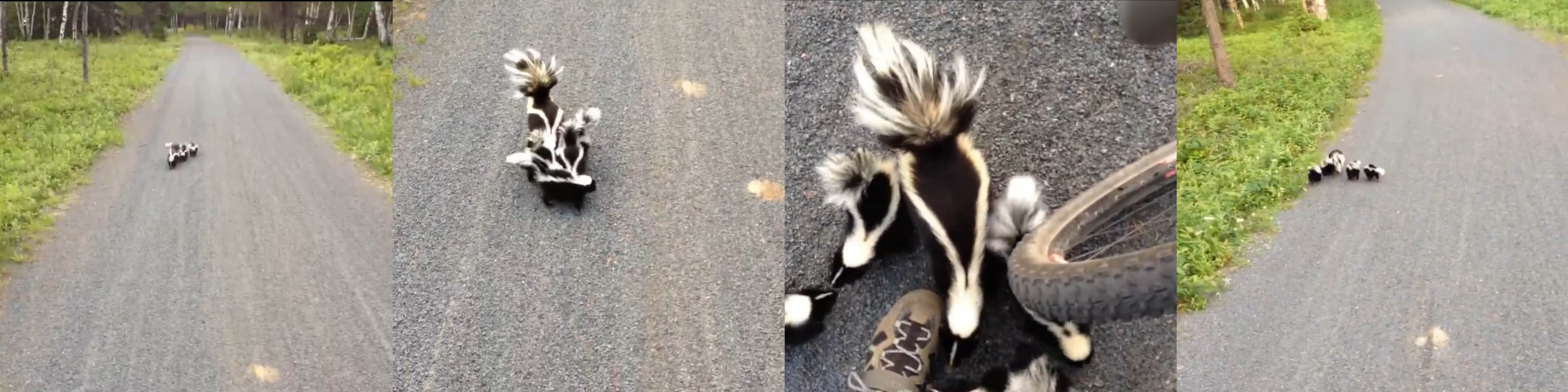}
}
\end{center}

\vspace{1mm}

\begin{tcolorbox}[examplesection, title=\textbf{Instruction}]
From the perspective of a person riding a bicycle on a rural gravel path, identify and describe the key events involving the animals encountered along the route. Summarize these events into a concise narrative, focusing on the encounter and interactions. Write the entire narrative in lowercase and end with a closing sentence that follows the pattern: ``thus ends the ride with [encounter description].''
\end{tcolorbox}

\vspace{1.2mm}

\begin{tcolorbox}[examplesection, colback=white, title=\textbf{Model Output}]
the journey begins with a solitary skunk appearing on the gravel path, its black and white fur contrasting against the grey surface. as the rider approaches, the skunk moves closer, its bushy tail held high. the encounter intensifies as the skunk's scent glands become visible, indicating a defensive posture. the rider slows down, and the skunk's movements become more erratic, suggesting a sense of alarm. the skunk's tail flares out, and it begins to move away, its body low to the ground. the rider stops, and the skunk pauses, its head turned towards the rider, creating a moment of tension. the skunk then scurries off the path, disappearing into the grassy verge. the rider remains stationary, watching the skunk retreat. the encounter concludes with the skunk's disappearance, leaving the rider alone on the path. thus ends the ride with a skunk's defensive display and subsequent retreat into the grass.
\end{tcolorbox}

\vspace{1.2mm}

\begin{tcolorbox}[checksection, title=\textbf{HYBRID CHECKS}]

\begin{tcolorbox}[constraintbox={\texttt{letter\_case}\hfill \Pass}]

\begin{tcolorbox}[evalitem]
\textbf{Function:}

{\ttfamily\small
check\_letter\_case(model\_output, case)
}

\vspace{0.35em}
\textbf{Extracted parameters:}
\begin{itemize}[leftmargin=1.5em, itemsep=0.15em, topsep=0.2em]
    \item \texttt{case = lowercase}
\end{itemize}

\vspace{0.2em}
\textbf{Result:} \Pass
\end{tcolorbox}

\end{tcolorbox}

\end{tcolorbox}

\end{tcolorbox}

\end{figure*}

\clearpage
\begin{figure*}[p]
\centering
\begin{tcolorbox}[exampleouter]

\begin{tcolorbox}[checksection, title=\textbf{LLM-BASED CHECKS}]

\begin{tcolorbox}[constraintbox={\texttt{summarize}\hfill \Fail}]

\begin{tcolorbox}[evalitem]
\textbf{Question 1:} Does the output mention the rider stop on a rural gravel path bordered by greenery and trees?

\vspace{0.35em}
\textbf{Result:} \Pass
\end{tcolorbox}

\vspace{0.45em}

\begin{tcolorbox}[evalitem]
\textbf{Question 2:} Does the output summarize the approach and interaction with the skunks, including their movement toward the rider and bicycle?

\vspace{0.35em}
\textbf{Result:} \Fail

\vspace{0.35em}
\textbf{Reason:}
\begin{itemize}[leftmargin=1.5em, itemsep=0.15em, topsep=0.2em]
    \item The output describes only a solitary skunk, while the intended event structure involves multiple skunks.
    \item It does not clearly summarize their approach and interaction with the rider and bicycle, and instead mainly emphasizes a defensive posture.
\end{itemize}
\end{tcolorbox}

\vspace{0.45em}

\begin{tcolorbox}[evalitem]
\textbf{Question 3:} Does the output mention the skunks eventually dispersing toward the edge of the path and into the vegetation?

\vspace{0.35em}
\textbf{Result:} \Pass
\end{tcolorbox}
\end{tcolorbox}

\vspace{0.7em}

\begin{tcolorbox}[constraintbox={\texttt{pattern\_closing}\hfill \Pass}]

\begin{tcolorbox}[evalitem]
\textbf{Constraint:} End with a closing sentence that follows the pattern: ``thus ends the ride with [encounter description].''

\vspace{0.35em}
\textbf{Result:} \Pass
\end{tcolorbox}

\end{tcolorbox}

\end{tcolorbox}

\end{tcolorbox}

\label{fig:appendix_example_skunk}
\end{figure*}

\begin{figure*}[p]
\centering

\begin{tcolorbox}[exampleouter]

\begin{tcolorbox}[exampletitle]
\textbf{\LARGE Example 4.}
\end{tcolorbox}

\vspace{1mm}

\begin{center}
\fcolorbox{cardborderA}{white}{
    \includegraphics[width=0.90\linewidth,height=4cm,keepaspectratio]{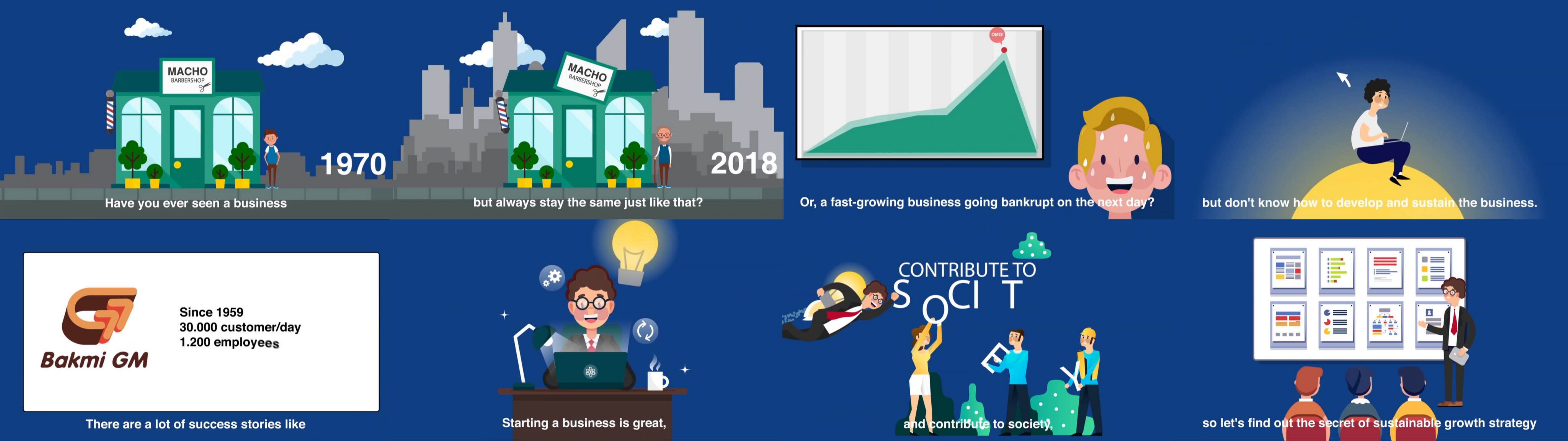}
}
\end{center}

\vspace{1mm}

\begin{tcolorbox}[examplesection, title=\textbf{Instruction}]
Present your prediction of the key factor that will enable a business to expand sustainably and positively impact society in the future as an unordered bullet list. Emphasize the forward-looking factor that ensures both growth and societal contribution, clearly linking the visible business patterns in the video to the predicted factor. Include at least one notable detail from the success story examples shown. Use an informative and professional tone, and avoid using contractions entirely.
\end{tcolorbox}

\vspace{1.2mm}

\begin{tcolorbox}[examplesection, colback=white, title=\textbf{Model Output}]
* \textbf{Deeply integrated, ethically governed artificial intelligence and adaptive learning systems} will be the pivotal factor enabling businesses to expand sustainably and positively impact society. This forward-looking factor signifies a fundamental shift from merely deploying technology for efficiency to embedding intelligence, continuous learning, and a robust ethical framework into the very core of business operations and strategic decision-making. For sustainable expansion, businesses will leverage AI to predict market shifts with unprecedented accuracy, optimize resource allocation to minimize waste across complex global supply chains, and personalize customer experiences while rigorously respecting privacy and data sovereignty. Simultaneously, ethical AI frameworks will ensure that these advancements contribute positively to society by promoting fairness, transparency, and accountability in algorithms, actively addressing complex global challenges such as climate change and healthcare disparities, and fostering inclusive economic growth. The ability to learn, adapt, and self-optimize with an inherent ethical compass will be paramount. Visible business patterns, such as the escalating investment in big data analytics, the proliferation of personalized marketing, the increasing demand for supply chain transparency, and the nascent but growing focus on explainable AI, are all precursors to this future state. Businesses are already collecting vast quantities of data and experimenting with AI, but the future demands a more holistic, ethical, and adaptive integration where AI systems continuously learn from their environment and societal feedback, driving both profitability and profound positive impact. A notable visual detail from successful future enterprises will be the vibrant, real-time digital dashboards prominently displayed in corporate lobbies, illustrating not only financial metrics but also environmental footprint reductions, community engagement scores, and ethical algorithm performance, serving as a transparent testament to their dual commitment to prosperity and planetary well-being.
\end{tcolorbox}

\end{tcolorbox}
\label{fig:appendix_example_prediction_page1}
\end{figure*}

\clearpage

\begin{figure*}[p]
\centering

\begin{tcolorbox}[exampleouter]

\begin{tcolorbox}[checksection, title=\textbf{RULE-BASED CHECKS}]

\begin{tcolorbox}[constraintbox={\texttt{bullet\_list}\hfill \Pass}]
\begin{tcolorbox}[evalitem]
\textbf{Function:}

{\ttfamily\small
check\_bullet\_list(model\_output)
}

\vspace{0.35em}
\textbf{Result:} \Pass
\end{tcolorbox}
\end{tcolorbox}

\vspace{0.7em}

\begin{tcolorbox}[constraintbox={\texttt{no\_contractions}\hfill \Pass}]
\begin{tcolorbox}[evalitem]
\textbf{Function:}

{\ttfamily\small
check\_no\_contractions(model\_output)
}

\vspace{0.35em}
\textbf{Constraint:} Avoid using contractions entirely.

\vspace{0.35em}
\textbf{Result:} \Pass
\end{tcolorbox}
\end{tcolorbox}

\end{tcolorbox}

\vspace{1.2mm}

\begin{tcolorbox}[checksection, title=\textbf{LLM-BASED CHECKS}]

\begin{tcolorbox}[constraintbox={\texttt{focus}\hfill \Pass}]
\begin{tcolorbox}[evalitem]
\textbf{Question 1:} Does the output primarily emphasize the predictive factor of a sustainable growth strategy as the enabler of future business expansion and societal contribution?

\vspace{0.35em}
\textbf{Result:} \Pass
\end{tcolorbox}
\end{tcolorbox}

\vspace{0.7em}

\begin{tcolorbox}[constraintbox={\texttt{content\_inclusion}\hfill \Fail}]
\begin{tcolorbox}[evalitem]
\textbf{Question 1:} Does the output explicitly mention at least one of the following notable visual details from the success story examples: (1) Bakmi GM's daily high customer volume and many employees, (2) Joger's transition from IDR500K to a top souvenir brand, or (3) Johnny Andrean's hundreds of outlets regionally?

\vspace{0.35em}
\textbf{Result:} \Fail

\vspace{0.35em}
\textbf{Reason:}
\begin{itemize}[leftmargin=1.5em, itemsep=0.15em, topsep=0.2em]
    \item The output introduces an invented futuristic detail about digital dashboards in corporate lobbies but fails to mention any of the required success-story details shown in the video.
\end{itemize}
\end{tcolorbox}
\end{tcolorbox}

\vspace{0.7em}

\begin{tcolorbox}[constraintbox={\texttt{tone}\hfill \Pass}]
\begin{tcolorbox}[evalitem]
\textbf{Constraint:} Use an informative and professional tone.

\vspace{0.35em}
\textbf{Result:} \Pass
\end{tcolorbox}
\end{tcolorbox}

\vspace{0.7em}

\begin{tcolorbox}[constraintbox={\texttt{causal}\hfill \Pass}]
\begin{tcolorbox}[evalitem]
\textbf{Constraint:} Clearly link the visible business patterns to the predicted factor.

\vspace{0.35em}
\textbf{Result:} \Pass
\end{tcolorbox}
\end{tcolorbox}

\end{tcolorbox}

\end{tcolorbox}
\label{fig:appendix_example_business_prediction}
\end{figure*}
\clearpage
\begin{figure*}[p]
\centering

\begin{tcolorbox}[exampleouter]

\begin{tcolorbox}[exampletitle]
\textbf{\LARGE Example 5.}
\end{tcolorbox}

\vspace{1mm}

\begin{center}
\fcolorbox{cardborderA}{white}{
    \includegraphics[width=0.90\linewidth,height=4cm,keepaspectratio]{figs/example5.pdf}
}
\end{center}

\vspace{1mm}

\begin{tcolorbox}[examplesection, title=\textbf{Instruction}]
First, observe the video and identify the visible cause-and-effect relationship between the chemical mixture and the resulting color transitions. Then, write an engaging promotional post for a science education platform. Ensure the whole output is valid JSON, does not include or imply any harmful or unsafe experimentation practices, uses at least one explicit sequence marker such as \texttt{first}, \texttt{then}, or \texttt{finally} to emphasize the progression, and clearly describes the causal relationship between the experiment setup and the color transformations.
\end{tcolorbox}

\vspace{1.2mm}

\begin{tcolorbox}[examplesection, colback=white, title=\textbf{Model Output}]
The promotional post should be informative, engaging, and suitable for social media platforms.

Here's the example of how the promotional post could look like in JSON format:

\texttt{json}

\{
  "post": \{
    "title": "Unlock the Magic of Color Chemistry!",
    "content": "Ever wondered what happens when you mix potassium permanganate with sugar and caustic soda? In our latest experiment, we observed a fascinating transformation from purple to green, then yellow! This isn't just about colors; it's about understanding chemical reactions and their effects on our environment. Join us on a journey to explore the wonders of science firsthand. \#ScienceForAll",
    "hashtags": ["\#ChemistryMagic", "\#ScienceExperiment"]
  \}
\}

\texttt{json}

\{
  "post": \{
    "title": "Discover the Wonders of Chemical Reactions!",
    "content": "Have you ever seen a chemical reaction change colors before your eyes? In this exciting experiment, we mixed potassium permanganate with sugar and caustic soda, first creating a vibrant purple solution. Then, as the reaction progressed, the solution transformed into a stunning green hue, finally settling into a bright yellow. This visual spectacle highlights the dynamic nature of chemistry and its ability to captivate our senses. Whether you're a seasoned scientist or a curious learner, there's always more to discover in the world of science. Dive in and let your imagination run wild! \#ExploreScience",
    "hashtags": ["\#ColorfulChemistry", "\#ScienceAdventure"]
  \}
\}
\end{tcolorbox}

\vspace{1.2mm}

\begin{tcolorbox}[checksection, title=\textbf{RULE-BASED CHECKS}]

\begin{tcolorbox}[constraintbox={\texttt{json}\hfill \Fail}]
\begin{tcolorbox}[evalitem]
\textbf{Function:}

{\ttfamily\small
check\_json(model\_output)
}

\vspace{0.35em}
\textbf{Constraint:} Ensure the output is valid JSON.

\vspace{0.35em}
\textbf{Result:} \Fail

\vspace{0.35em}
\textbf{Reason:}
\begin{itemize}[leftmargin=1.5em, itemsep=0.15em, topsep=0.2em]
    \item The output is not a single valid JSON object. It includes two separate JSON blocks while our evaluation extracts the substring from the first \texttt{\{} to the last \texttt{\}} and attempts to parse it as one complete JSON object.
\end{itemize}
\end{tcolorbox}
\end{tcolorbox}

\end{tcolorbox}

\end{tcolorbox}
\end{figure*}

\clearpage

\begin{figure*}[p]
\centering

\begin{tcolorbox}[checksection, title=\textbf{LLM-BASED CHECKS}]

\begin{tcolorbox}[constraintbox={\texttt{harmful\_content\_exclusion}\hfill \Pass}]
\begin{tcolorbox}[evalitem]
\textbf{Constraint:} Do not include or imply any harmful or unsafe experimentation practices.

\vspace{0.35em}
\textbf{Result:} \Pass

\vspace{0.35em}
\end{tcolorbox}
\end{tcolorbox}

\vspace{0.7em}

\begin{tcolorbox}[constraintbox={\texttt{sequence\_marker\_inclusion}\hfill \Pass}]
\begin{tcolorbox}[evalitem]
\textbf{Constraint:} Use at least one explicit sequence marker such as \texttt{first}, \texttt{then}, or \texttt{finally} to emphasize the progression.

\vspace{0.35em}
\textbf{Result:} \Pass
\end{tcolorbox}
\end{tcolorbox}

\vspace{0.7em}

\begin{tcolorbox}[constraintbox={\texttt{causal}\hfill \Pass}]
\begin{tcolorbox}[evalitem]
\textbf{Constraint:} Clearly describe the causal relationship between the experiment setup and the color transformations.

\vspace{0.35em}
\textbf{Result:} \Pass
\end{tcolorbox}
\end{tcolorbox}

\end{tcolorbox}

\label{fig:appendix_example_science_v28_case2}
\end{figure*}
\FloatBarrier
\clearpage
\FloatBarrier
\clearpage

\begin{figure*}[t]
\centering
\begin{tcolorbox}[
    width=0.96\textwidth,
    colback=gray!8,
    colframe=black!55,
    boxrule=0.7pt,
    arc=2mm,
    left=2.5mm,
    right=2.5mm,
    top=1.5mm,
    bottom=1.5mm
]
\begin{lstlisting}[basicstyle=\ttfamily\scriptsize, breaklines=true, columns=fullflexible]
You are an evaluation assistant.

We will validate ONE constraint using a LOCAL Python function.
You must decide which parameters to pass into the function based on:
- constraint_content (natural language constraint)
- function_signature (what arguments exist)
- param_docs (meaning of each parameter)
- the model_output (the candidate answer)

CONSTRAINT:
- constraint_id: {constraint_id}
- constraint_content: {constraint_content}

LOCAL FUNCTION:
- function_id: {function_id}
- function_signature: {function_signature}
- usage: {usage}
- param_docs: {json.dumps(param_docs, ensure_ascii=False)}

MODEL OUTPUT (for reference only; do NOT repeat it back):
<<<MODEL_OUTPUT_START
{model_output}
MODEL_OUTPUT_END>>>

Your task:
1) Determine which parameters (excluding model_output) are needed to evaluate this constraint.
2) Produce parameter values ONLY, separated by the exact delimiter: |||
3) The first argument is ALWAYS model_output and is implicit; you MUST NOT output it.
4) Prefer key=value format for safety, especially for optional parameters.
5) For List values, output JSON arrays; for numbers output plain numbers.
6) Output must be a SINGLE LINE with no extra text.

Now output the parameters line:
\end{lstlisting}
\end{tcolorbox}
\caption{Prompt used for \textbf{hybrid verification}. Given a natural-language constraint, a local checking function, and its parameter documentation, the model extracts the required function arguments so that the final checker can be executed programmatically.}
\label{fig:hybrid_eval_prompt}
\end{figure*}

\begin{figure*}[t]
\centering
\begin{tcolorbox}[
    width=0.96\textwidth,
    colback=gray!8,
    colframe=black!55,
    boxrule=0.7pt,
    arc=2mm,
    left=2.5mm,
    right=2.5mm,
    top=1.5mm,
    bottom=1.5mm
]
\begin{lstlisting}[basicstyle=\ttfamily\scriptsize, breaklines=true, columns=fullflexible]
You are a strict multimodal judge. You can see the video frames attached after this text.

TASK:
For EACH constraint_id in the list, decide whether the CANDIDATE ANSWER satisfies the constraint, given:
- VIDEO FRAMES (visual evidence)
- ORIGINAL INSTRUCTION
- constraint_content

OUTPUT RULES (EXTREMELY IMPORTANT):
1) Output MUST be a single valid JSON object ONLY. No markdown, no extra text.
2) You MUST include EVERY constraint_id key listed below.
3) Each value MUST be exactly 0 or 1 (integer). NOT true/false.
4) If unsure, output 0 for that constraint_id.
5) Keys must match the given constraint_id strings exactly.

CONSTRAINT_ID LIST (MUST respond to ALL of them):
{json.dumps(cids, ensure_ascii=False)}

ORIGINAL INSTRUCTION:
<<<INSTRUCTION_START
{instruction}
INSTRUCTION_END>>>

CANDIDATE ANSWER (MODEL OUTPUT):
<<<MODEL_OUTPUT_START
{model_output}
MODEL_OUTPUT_END>>>

CONSTRAINTS (id + content):
{json.dumps(constraints_pack, ensure_ascii=False, indent=2)}

Now output ONLY the JSON object:
\end{lstlisting}
\end{tcolorbox}
\caption{Prompt used for \textbf{LLM-based judging} of structure-related constraints. Given the video frames, original instruction, candidate answer, and constraint descriptions, the judge model outputs a JSON object containing binary decisions for each constraint.}
\label{fig:structure_judge_prompt}
\end{figure*}

\begin{figure*}[t]
\centering
\begin{tcolorbox}[
    width=0.96\textwidth,
    colback=gray!8,
    colframe=black!55,
    boxrule=0.7pt,
    arc=2mm,
    left=2.5mm,
    right=2.5mm,
    top=1.5mm,
    bottom=1.5mm
]
\begin{lstlisting}[basicstyle=\ttfamily\scriptsize, breaklines=true, columns=fullflexible]
You are a strict judge.

Decide whether the MODEL OUTPUT satisfies the SUB-CONSTRAINT question.

CONSTRAINT:
- constraint_id: {constraint_id}
- constraint_content: {constraint_content}

SUB-CONSTRAINT QUESTION:
{sub_q}

{inst_block}

MODEL OUTPUT:
<<<MODEL_OUTPUT_START
{model_output}
MODEL_OUTPUT_END>>>

Answer with exactly one token: YES or NO.
If unsure, answer NO.
\end{lstlisting}
\end{tcolorbox}
\caption{Prompt used for \textbf{LLM-based judging} of style-related, format-related, and selected content-related constraints. For some content-related constraints, including \texttt{content\_inclusion}, \texttt{tracking}, \texttt{summarize}, \texttt{focus}, and \texttt{selection\_accuracy}, we decompose each high-level constraint into manually written sub-constraint questions (\texttt{sub\_q}) to support more explicit and fine-grained judgment.}
\label{fig:subq_judge_prompt}
\end{figure*}

\FloatBarrier
\FloatBarrier
\clearpage
\FloatBarrier
\clearpage

\begin{figure*}[t]
\centering
\begin{tcolorbox}[
    width=0.96\textwidth,
    colback=gray!8,
    colframe=black!55,
    boxrule=0.7pt,
    arc=2mm,
    left=2.5mm,
    right=2.5mm,
    top=1.5mm,
    bottom=1.5mm
]
\begin{lstlisting}[basicstyle=\ttfamily\scriptsize, breaklines=true, columns=fullflexible]
You are an expert dataset item writer for a VIDEO-ONLY instruction-following benchmark.

MODE-SPECIFIC RULES
You are given an MCQ item (question, options, and answer for verification only).

You MUST:
- Rewrite it as an OPEN-ENDED instruction.
- Remove all multiple-choice artifacts, including option labels, explicit answer choices, and answer hints.
- Preserve the original semantic target while making the instruction natural and answerable from video frames.
- If the item is counterfactual or predictive, ensure that the instruction depends on video-specific observable cues and cannot be answered by generic reasoning alone.

INTRINSIC REQUIREMENTS
- The instruction must not be solvable by generic commonsense alone.

ATOMICITY AND VERIFIABILITY
- Default: generate ONE atomic core task only.
- Do NOT merge multiple unrelated semantic goals.
- Every requirement in the instruction and every bullet under "Additional constraints:" must be decidable from video frames.
- Do NOT ask for anything that cannot be inferred from the frames.
- The instruction may contain up to THREE steps, but all steps must remain within the same semantic task.

COMPOSITION GUIDANCE
- Use "And" for the default single-output task.
- Use "Chain" only when the task cleanly decomposes into multiple frame-grounded steps.
- Use "Selection" only when the branch condition is decidable from video frames and the selected branch should be explicitly recorded in the field "Selection".

SELECTION RULES
- A Selection instruction must include:
  - "Condition: <question decidable from frames>"
  - "Options:" with at least two stable labels
  - "Branches:" with "If <OptionLabel>: <requirements>" for every option
- The output field "Selection" must exactly match one option label.
- The "Additional constraints:" bullets must describe only the selected branch.

CONSTRAINT RULES
- Each constraint bullet must be relevant, checkable, and aligned one-to-one with constraint_dimensions.
- Do NOT add trivial bullets such as "do not speculate beyond what is visible" or "do not mention any on-screen text".

{min_constraints_rule_block}

OUTPUT REQUIREMENTS
- Choose exactly ONE task_type from the predefined task taxonomy.
- Do NOT invent new task types or new constraint IDs outside the given taxonomy.

GLOSSARY
{task_type_glossary}

{constraint_id_glossary}

OUTPUT FORMAT (STRICT JSON; no markdown fences)

Return exactly ONE JSON object in the following unified format:

{
  "task_type": "<one id from TASK TAXONOMY>",
  "composition_type": "And" | "Chain" | "Selection" | ["And" | "Chain" | "Selection", ...],
  "instruction": "<instruction text>",
  "constraint_dimensions": ["<constraint_id 1>", "<constraint_id 2>", ...],
  "reference_answer": "<evaluation-only reference>",
  "Selection": "<selected option label decided from frames; include this field ONLY when composition_type contains Selection>"
}

CONSTRAINT TAXONOMY IDS
{json.dumps(CONSTRAINT_IDS, ensure_ascii=False, indent=2)}

INPUT (JSON) --- The answering model will NOT see this:
{payload_str}
\end{lstlisting}
\end{tcolorbox}
\caption{Prompt used for \textbf{MCQ-based instruction generation}. Starting from an existing multiple-choice item, the generator rewrites it into an open-ended, video-grounded instruction while enforcing atomicity, frame-grounded verifiability, composition validity, and constraint quality.}
\label{fig:mcq_generation_prompt}
\end{figure*}

\begin{figure*}[t]
\centering
\begin{tcolorbox}[
    width=0.96\textwidth,
    colback=gray!8,
    colframe=black!55,
    boxrule=0.7pt,
    arc=2mm,
    left=2.5mm,
    right=2.5mm,
    top=1.5mm,
    bottom=1.5mm
]
\begin{lstlisting}[basicstyle=\ttfamily\scriptsize, breaklines=true, columns=fullflexible]
You are an expert dataset item writer for a VIDEO-ONLY instruction-following benchmark.

MODE-SPECIFIC RULES
You are given a forced task_type, composition_type, and constraint_dimensions.

You should follow them as closely as possible. However, you may:
- ADD extra constraint_id(s), or
- REPLACE a constraint_id

only when this is truly necessary to make the item coherent, answerable from video frames, and verifiable.

If you add or replace anything, you MUST briefly state it in the AUXILIARY section:
- what was changed (added/replaced), and
- why the change was necessary.

INTRINSIC REQUIREMENTS
- The instruction must not be solvable by generic commonsense alone.
- If the task is counterfactual or predictive, it must be grounded in video-specific observable cues.

ATOMICITY AND VERIFIABILITY
- Default: generate ONE atomic core task only.
- Do NOT merge multiple unrelated semantic goals.
- Every requirement in the instruction and every bullet under "Additional constraints:" must be decidable from video frames.
- Do NOT ask for anything that cannot be inferred from the frames.
- The instruction may contain up to THREE steps, but all steps must remain within the same semantic task.

COMPOSITION GUIDANCE
- Use "And" for the default single-output task.
- Use "Chain" only when the task cleanly decomposes into multiple frame-grounded steps.
- Use "Selection" only when the branch condition is decidable from video frames and the selected branch should be explicitly recorded in the field "Selection".

SELECTION RULES
- A Selection instruction must include:
  - "Condition: <question decidable from frames>"
  - "Options:" with at least two stable labels
  - "Branches:" with "If <OptionLabel>: <requirements>" for every option
- The output field "Selection" must exactly match one option label.
- The "Additional constraints:" bullets must describe only the selected branch.

CONSTRAINT RULES
- Each constraint bullet must be relevant, checkable, and aligned one-to-one with constraint_dimensions.
- Do NOT add trivial bullets such as "do not speculate beyond what is visible" or "do not mention any on-screen text".

{min_constraints_rule_block}

OUTPUT REQUIREMENTS
- Choose exactly ONE task_type from the predefined task taxonomy.
- Do NOT invent new task types or new constraint IDs outside the given taxonomy.

GLOSSARY
{task_type_glossary}

{constraint_id_glossary}

OUTPUT FORMAT (STRICT JSON; no markdown fences)

Return exactly ONE JSON object in the following unified format:

{
  "task_type": "<one id from TASK TAXONOMY>",
  "composition_type": "And" | "Chain" | "Selection" | ["And" | "Chain" | "Selection", ...],
  "instruction": "<instruction text>",
  "constraint_dimensions": ["<constraint_id 1>", "<constraint_id 2>", ...],
  "reference_answer": "<evaluation-only reference>",
  "Selection": "<selected option label decided from frames; include this field ONLY when composition_type contains Selection>"
}

CONSTRAINT TAXONOMY IDS
{json.dumps(CONSTRAINT_IDS, ensure_ascii=False, indent=2)}

INPUT (JSON) --- The answering model will NOT see this:
{payload_str}
\end{lstlisting}
\end{tcolorbox}
\caption{Prompt used for \textbf{video-grounded instruction generation}. Given forced task, composition, and constraint settings together with video-derived input, the generator constructs a benchmark item under rules enforcing atomicity, frame-grounded verifiability, composition validity, and constraint quality.}
\label{fig:video_generation_prompt}
\end{figure*}

\FloatBarrier

\begin{figure*}[t]
\centering
\begin{tcolorbox}[
    width=0.96\textwidth,
    colback=gray!8,
    colframe=black!55,
    boxrule=0.7pt,
    arc=2mm,
    left=2.5mm,
    right=2.5mm,
    top=1.5mm,
    bottom=1.5mm
]
\begin{lstlisting}[basicstyle=\ttfamily\scriptsize, breaklines=true, columns=fullflexible]
You are rewriting ONE instruction-following annotation into a clean, executable form for evaluation.

You are given ONE record containing:
- captions: auxiliary text for the item writer/evaluator
- instruction: the instruction shown to the answering model
- constraint_dimensions: a list of constraint_id tokens (which may contain errors)
- CONSTRAINT_ID_DEFINITION: the ONLY valid constraint_id keys and their meanings

Composition types:
- And: satisfy multiple constraints at the same time.
- Chain: complete multiple tasks in sequence, where later tasks may depend on earlier outputs.
- Selection: evaluate a condition and follow the corresponding branch.

Your job:
1) Extract constraints from any constraint-list section (e.g., "Additional constraints:"), regardless of bullet style.
   Treat each bullet line as one constraint candidate, even if it contains ":" or ";" internally.

2) Rewrite everything into ONE natural instruction string called "normalized_instruction":
   - remove headings such as "Additional constraints:"
   - remove bullet markers or numbering
   - merge all requirements into one coherent instruction

3) If composition_type is "Selection", rewrite normalized_instruction in fluent ENGLISH using this structure:
   - first, write 1--2 sentences introducing the task, condition, and options
   - then, write one paragraph in the following form:
     "If Option A: <task requirement + Option A constraints>. If Option B: <task requirement + Option B constraints>."

4) Align and clean up constraint IDs:
   - map each extracted constraint to one or more IDs from CONSTRAINT_ID_DEFINITION
   - if multiple IDs apply, output "constraint_id" as a list
   - correct misspelled IDs in constraint_dimensions by mapping them to the closest valid key
   - add missing IDs only when they are directly supported by the extracted constraint text
   - delete extra or invalid original IDs

5) Perform a final consistency check:
   - detect conflicts or non-executable constraints
   - revise them minimally to make them executable and non-conflicting
   - record each change in "revisions" with a short reason

Output STRICT JSON ONLY:

{
  "normalized_instruction": "<ONE merged natural instruction>",
  "composition_type": "And" | "Chain" | "Selection",
  "constraints": [
    {
      "constraint_id": "<a key OR a list of keys from CONSTRAINT_ID_DEFINITION>",
      "constraint_content": "<constraint clause>"
    }
  ],
  "fixed_constraint_dimensions": ["<ids after correction>"],
  "revisions": [{"type":"...", "before":"...", "after":"...", "reason":"..."}],
  "warnings": ["<optional>"]
}

Rules:
- constraint_id must use ONLY keys from CONSTRAINT_ID_DEFINITION
- output STRICT JSON only

CONSTRAINT_ID_DEFINITION:
{constraint_def_str}

INPUT (JSON):
{payload_str}
\end{lstlisting}
\end{tcolorbox}
\caption{Prompt used for \textbf{annotation normalization and constraint repair}. Given an instruction-following annotation record, the model rewrites it into an executable normalized instruction, aligns extracted constraints with valid taxonomy IDs, and performs minimal revisions to resolve conflicts or non-executable requirements.}
\label{fig:annotation_normalization_prompt}
\end{figure*}

\begin{figure*}[t]
\centering
\begin{tcolorbox}[
    width=0.96\textwidth,
    colback=gray!8,
    colframe=black!55,
    boxrule=0.7pt,
    arc=2mm,
    left=2.5mm,
    right=2.5mm,
    top=1.5mm,
    bottom=1.5mm
]
\begin{lstlisting}[basicstyle=\ttfamily\scriptsize, breaklines=true, columns=fullflexible]
You are evaluating whether the CANDIDATE ANSWER recognizes and localizes
incompatible requirements in the instruction.

Judge only what the candidate explicitly states or unambiguously paraphrases.
Do not infer that the candidate noticed a conflict merely because the
annotation below reveals one. Use EXPECTED MINIMAL CORE only to map the
requirements mentioned by the candidate to their UIDs.

Return exactly one JSON object:
{
  "conflict_detected": 0 or 1,
  "localized_core_uids": ["UID", "..."]
}

Rules:
- conflict_detected=1 only if the candidate clearly states that two or more
  requirements conflict or cannot all be satisfied as written.
  Difficulty, uncertainty, missing information, and unrelated refusals do not
  qualify.
- localized_core_uids may contain only UIDs from EXPECTED MINIMAL CORE.
  Include a UID only when the candidate explicitly identifies, quotes, or
  unambiguously paraphrases that requirement as part of the conflict.
- Do not infer requirements that the candidate does not mention. A generic
  statement such as "the constraints conflict" detects the conflict but
  localizes no UID, so return an empty list.
- For complete localization, every UID in EXPECTED MINIMAL CORE must be
  identified. This rule applies equally to two-way and three-way conflicts.

CONFLICT ID: {conflict_id}

EXPECTED MINIMAL CORE:
{expected_minimal_core_json}

INSTRUCTION:
<<<INSTRUCTION_START
{instruction}
INSTRUCTION_END>>>
CANDIDATE ANSWER:
<<<CANDIDATE_START
{candidate_answer}
CANDIDATE_END>>>

Output only the JSON object.
\end{lstlisting}
\end{tcolorbox}
\caption{Prompt used for \textbf{Conflict-100 evaluation}.}
\label{fig:conflict_judge_prompt}
\end{figure*}
\end{document}